\documentclass{article} 
\usepackage{iclr2015,times}
\usepackage{hyperref}
\usepackage{graphicx,graphics,wrapfig,epstopdf,subfig}
\usepackage{graphicx}
\usepackage{amsmath}
\usepackage{amssymb}
\usepackage{algorithm,algorithmic}
\usepackage{amsmath,amsthm,amsfonts,amssymb,amscd}
\usepackage{fullpage}
\usepackage{lastpage}
\usepackage{enumerate}
\usepackage{fancyhdr}
\usepackage{mathrsfs}
\usepackage{xcolor}
\usepackage{graphicx}
\usepackage{bbm}
\usepackage{float}
\usepackage{booktabs}
\usepackage{bm}
\usepackage{amsmath,amssymb} 
\newtheorem{lemma}{Lemma} 
\newcommand{\Dmat}{{\bf D}}

\newcommand{\Xmat}{{\bf X}}

\newcommand{\thetav}{\boldsymbol{\theta}}

\iclrconference

\title{Generative Deep Deconvolutional Learning}

\author{
Yunchen Pu, Xin Yuan and Lawrence Carin \\
Department of Electrical and Computer Engineering, Duke University,
Durham, NC, 27708, USA \\
\texttt{\{yunchen.pu,xin.yuan,lcarin\}@duke.edu} 
}
\begin{document}

\maketitle

\begin{abstract}
A generative model is developed for deep (multi-layered) convolutional dictionary learning.
A novel {probabilistic pooling} operation is integrated into the deep model, yielding efficient bottom-up (pretraining) and top-down (refinement) probabilistic learning. After learning the deep convolutional dictionary, testing is implemented via deconvolutional inference. To speed up this inference, a new statistical approach is proposed to project the top-layer dictionary elements to the data level. Following this, 
only one layer of deconvolution is required during testing.
Experimental results demonstrate powerful capabilities of the model to learn multi-layer features from images, and excellent classification results are obtained on the MNIST and Caltech 101 datasets. 
\end{abstract}

\section{Introduction}
Convolutional networks, introduced in~\cite{Lecun98gradient-basedlearning}, have demonstrated excellent performance on image classification and other tasks. 
There are at least two key components of this model: computational efficiency manifested by leveraging the convolution operator, and a deep architecture, in which the features of a given layer serve as the inputs to the next layer above.
Since that seminal contribution, much work has been undertaken on improving deep convolutional networks~\citep{Lecun98gradient-basedlearning},
deep deconvolutional networks~\citep{Zeiler10CVPR},
convolutional deep restricted Boltzmann machines~\citep{Lee09ICML}, 
and on Bayesian convolutional dictionary learning~\citep{Chen13deepCFA}, among others.

An important technique employed in these deep models is {\em pooling}, in which a contiguous block of features from the layer below are mapped to a single input feature for the layer above. The pooling step manifests robustness, by minimizing the effects of variations due to small shifts, and it has the advantage of reducing the number of features as one moves higher in the hierarchical representation (possibly mitigating over-fitting). 
Methods that have been considered include average and maximum pooling, in which the single feature mapped as input to the layer above is respectively the average or maximum of the corresponding block of features below. Average pooling may introduce blur to learned filters~\citep{Zeiler13ICLR}, and use of the maximum (``max pooling'') is widely employed. Note that average and max pooling are deterministic. 
Stochastic pooling proposed by~\cite{Zeiler13ICLR} and the probabilistic max-pooling used by~\cite{Lee09ICML} often improve the pooling process. The use of stochastic pooling is also attractive in the context of developing a generative model for the deep convolutional representation, as highlighted in this paper. Specifically, we develop a deep generative statistical model, which starts at the highest-level features, and maps these through a sequence of layers, until ultimately mapping to the data plane (e.g., an image). The feature at a given layer is mapped via a multinomial distribution to one feature in a block of features at the layer below (and all other features in the block at the next layer are set to zero). 
This is analogous to the method in~\cite{Lee09ICML}, in the sense of imposing that there is {\em at most} one non-zero activation within a pooling block. 
As we demonstrate, this yields a generative statistical model with which Bayesian inference may be readily implemented, with all layers analyzed jointly to fit the data. 

We use bottom-up pretraining, in which initially we sequentially learn parameters of each layer one at a time, from bottom to top, based on the features at the layer below. However, in the refinement phase, all model parameters are learned jointly, top-down. Each consecutive layer in the model is locally conjugate in a statistical sense, so learning model parameters may be readily performed using sampling or variational methods. We here develop a Gibbs sampler for learning, with the goal of obtaining a maximum \emph{a posterior} (MAP) estimate of the model parameters, as in the original paper on Gibbs sampling \citep{Geman1984} (we have found it unnecessary, and too expensive, to attempt an accurate estimate of the full posterior). The Gibbs sampler employed for parameter learning may be viewed as an alternative to typical optimization-based learning \citep{Lecun98gradient-basedlearning,Zeiler10CVPR}, making convenient use of the developed generative statistical model. 

The work in \cite{Zeiler10CVPR,Chen13deepCFA} involves learning convolutional dictionaries, and at the testing phase one must perform a (generally) expensive nonlinear deconvolution step at each layer. In \cite{LeCun10NIPS} convolutional dictionaries are also learned at the training stage, but one also simultaneously learns a convolutional filterbank and nonlinear function. The convolutional filterbank can be implemented quickly at test (no nonlinear deconvolutional inversion) and, linked with the nonlinear function, this computationally efficient testing step is meant to approximate the devonvolutional network. 

We propose an alternative approach to yield fast inversion at test, while still retaining an aspect of the nonlinear deconvolution operation. As detailed below, in the learning phase, we infer a deep hierarchy of convolutional dictionary elements, which if handled like in \cite{Zeiler10CVPR}, requires joint deconvolution at each layer when testing. However, leveraging our generative statistical model, the dictionary elements at the top of the hierarchy can be mapped through a sequence of linear operations to the image/data plane. At test, we only employ the features from the top layer in the hierarchy, mapped to the data plane, and therefore only a single layer of deconvolution need be applied. This implies that the test-time computational cost is independent of the number of layers employed during the learning phase. 
 
This paper makes three contributions: ($i$) rather than employing beta-Bernoulli sparsity at each layer of the model separately, as in \cite{ICML2011Chen,Chen13deepCFA}, the sparsity is manifested via a multinomial process between layers, constituting stochastic pooling, and allowing coupling all layers of the deep model when learning; ($ii$) the stochastic pooling manifests a proper top-down generative model, allowing a new means of mapping high-level features to the data plane; and ($iii$) a novel form of testing is employed with deep models, with the top-layer features mapped to the data plane, and deconvolution only applied once, directly with the data. This methodology yields excellent performance on image-recognition tasks, as demonstrated in the experiments.

\vspace{-3mm}
\section{Modeling Framework}
\vspace{-2mm}
The proposed model is applicable to general data for which a convolutional dictionary representation is appropriate. One may, for example, apply the model to one-dimensional signals such as audio, or to two-dimensional imagery. In this paper we focus on imagery, and hence assume two-dimensional signals and convolutions. Gray-scale images are considered for simplicity, with straightforward extension to color.

\vspace{-3mm}
\subsection{Single-Layer Convolutional Dictionary Learning}
\vspace{-2mm}
Assume $N$ gray-scale images $\{{\Xmat^{(n)}}\}_{n=1,N}$, with $\Xmat^{(n)}\in\mathbb{R}^{N_x \times N_y}$; the images are analyzed jointly to learn the convolutional dictionary $\{{\Dmat^{(k)}}\}_{k=1,K}$. Specifically
consider the model 
\vspace{-2mm}
\begin{equation}\label{Eq:betabern}
\Xmat^{(n)} = \sum_{k=1}^K  \Dmat^{(k)} \ast ({\bf Z}^{(n,k)}\odot {\bf W}^{(n,k)})  + {\bf E}^{(n)},
\vspace{-2mm}
\end{equation}
where $\ast$ is the convolution operator, $\odot$ denotes the Hadamard (element-wise) product, the elements of ${\bf Z}^{(n,k)}$ are in $\{0,1\}$, the elements of ${\bf W}^{(n,k)}$ are real, and ${\bf E}^{(n)}$ represents the residual. ${\bf Z}^{(n,k)}$ indicates which shifted version of ${\bf D}^{(k)}$ is used to represent ${\bf X}^{(n)}$. 
Considering ${\Dmat}^{(k)}\in {\mathbb R}^{n_{d_x} \times n_{d_y}}$ (typically $n_{d_x}\ll N_x$ and $n_{d_y}\ll N_y$), the corresponding weights ${\bf Z}^{(n,k)}\odot {\bf W}^{(n,k)}$ are of size $(N_x-n_{d_x} +1)\times(N_y - n_{d_y} +1)$.

Let $w_{i,j}^{(n,k)}$ and $z_{i,j}^{(n,k)}$ represent elements $(i,j)$ of ${\bf Z}^{(n,k)}$ and ${\bf W}^{(n,k)}$, respectively. Within a Bayesian construction, the priors for the model may be represented as 
\citep{Paisley09ICML}:
\begin{eqnarray}
{ z}^{(n,k)}_{i,j} &\sim& {\rm Bernoulli}(\pi_{i,j}^{(n,k)}), \quad \quad \quad \pi_{i,j}^{(n,k)} \sim {\rm Beta}(a_0, b_0), \label{eq:beta-bern}\\
{ w}^{(n,k)}_{i,j} &\sim& {\cal N}(0, \gamma_w^{-1}), ~~\quad {\bf D}^{(k)}\sim {\cal N}(0, \gamma_d^{-1}{\bf I}), \quad {\bf E}^{(n)}\sim {\cal N}(0, \gamma_e^{-1}{\bf I}), \\
\gamma_w &\sim& {\rm Ga}(a_w, b_w),  \quad\quad \gamma_d \sim {\rm Ga}(a_d, b_d), \quad\quad~~ \gamma_e \sim {\rm Ga}(a_e, b_e), 
\end{eqnarray}
where $i= 1,\dots, N_x-n_{d_x} +1;~~j=1,\dots, N_y - n_{d_y} +1$, ${\rm Ga(\cdot)}$ denotes the gamma distribution, ${\bf I}$ represents the identity matrix, and $\{a_0,b_0, a_w,b_w, a_d, b_d, a_e, b_e\}$ are hyperparameters, for which default settings are discussed in \cite{Paisley09ICML,ICML2011Chen,Chen13deepCFA}. While the model may look somewhat complicated, local conjugacy admits Gibbs sampling or variational Bayes inference \citep{ICML2011Chen,Chen13deepCFA}.

In \cite{ICML2011Chen,Chen13deepCFA} a deep model was developed based on (\ref{Eq:betabern}), by using ${\bf  S}^{(n,k)}\stackrel{\rm def}{=}{\bf  Z}^{(n,k)}\odot{\bf  W}^{(n,k)}$ as the input of the layer above. In order to do this, a pooling operation ({\em e.g.}, the max-pooling used in~\cite{Chen13deepCFA}) is employed, reducing the feature dimension as one moves to higher layers.
However, the model was learned by stacking layers upon each other, without subsequent overall refinement. This was because use of deterministic max pooling undermined development of a proper top-down generative model that coupled all layers; therefore, in \cite{Chen13deepCFA} the model in (\ref{Eq:betabern}) was used sequentially from bottom-up, but the overall model parameters were never coupled when learning.
To tackle this, we propose a {\em probabilistic pooling} procedure, yielding a top-down deep generative statistical structure, coupling all parameters when performing learning. As discussed when presenting results, this joint learning of all layers plays a critical role in improving model performance. The stochastic pooling applied here is closely related to that in \cite{Zeiler13ICLR,Lee09ICML}.

\vspace{-3mm}
\subsection{Pretraining \& Stochastic Pooling}
\vspace{-3mm}

Parameters of the deep model are learned by first analyzing one layer of the model at a time, starting at the bottom layer (touching the data), and sequentially stacking layers. The parameters of each layer of the model are learned separately, conditioned on parameters of the layers learned thus far (like in \cite{ICML2011Chen,Chen13deepCFA}). The parameters learned in this manner serve as {\em initializations} for the top-down refinement step, discussed in Sec.~\ref{sec:refine}, in which parameters at all layers of the deep model are learned jointly.

Assume an $L$-layer model, with layer $L$ the top layer, and layer 1 at the bottom, closest to the data. In the pretraining stage, the output of layer $l$ is the input to layer $l+1$, after pooling. 
Layer $l\in\{1,\dots,L\}$ has $K_l$ dictionary elements, and we have:
\begin{eqnarray}
{\bf X}^{(n, l+1)} &=& \sum_{k_{l+1}=1}^{K_{l+1}} {\bf D}^{(k_{l+1}, l+1)} * \left({\bf Z}^{(n,k_{l+1},l+1)}  \odot {\bf W}^{(n,k_{l+1}, l+1)}\right) + {\bf E}^{(n, l+1)} \label{Eq:x_lp1}\\
{\bf X}^{(n, l)} &=& \sum_{k_{l}=1}^{K_{l}} {\bf D}^{(k_{l}, l)} * \underbrace{\left({\bf Z}^{(n,k_{l},l)} \odot {\bf W}^{(n,k_{l}, l)}\right)}_{= {\bf S}^{(n,k_{l},l)}} + {\bf E}^{(n, l)} \label{Eq:x_l}
\end{eqnarray}
The expression ${\bf S}^{(n,k_l,l)}$ is a 2D (spatial) activation map, for image $n$, model layer $l$, dictionary element $k_l$. The expression ${\bf X}^{(n,l+1)}$ may be viewed as a 3D entity, with its $k_l$-th plane defined by a ``pooled'' version of ${\bf S}^{(n,k_l,l)}$ (pooling discussed next). The dictionary elements ${\bf D}^{(k_l,l)}$ and residual ${\bf E}^{(n,l)}$ are also three dimensional (each 2D plane of ${\bf D}^{(k_l,l)}$ and ${\bf E}^{(n,l)}$ is the spatial-dependent structure of the corresponding features), and the convolution is performed in the 2D spatial domain, simultaneously for each layer of the feature map.

\begin{figure}[tbp!]
	\centering
	\vspace{-3mm}
	\includegraphics[scale=0.45]{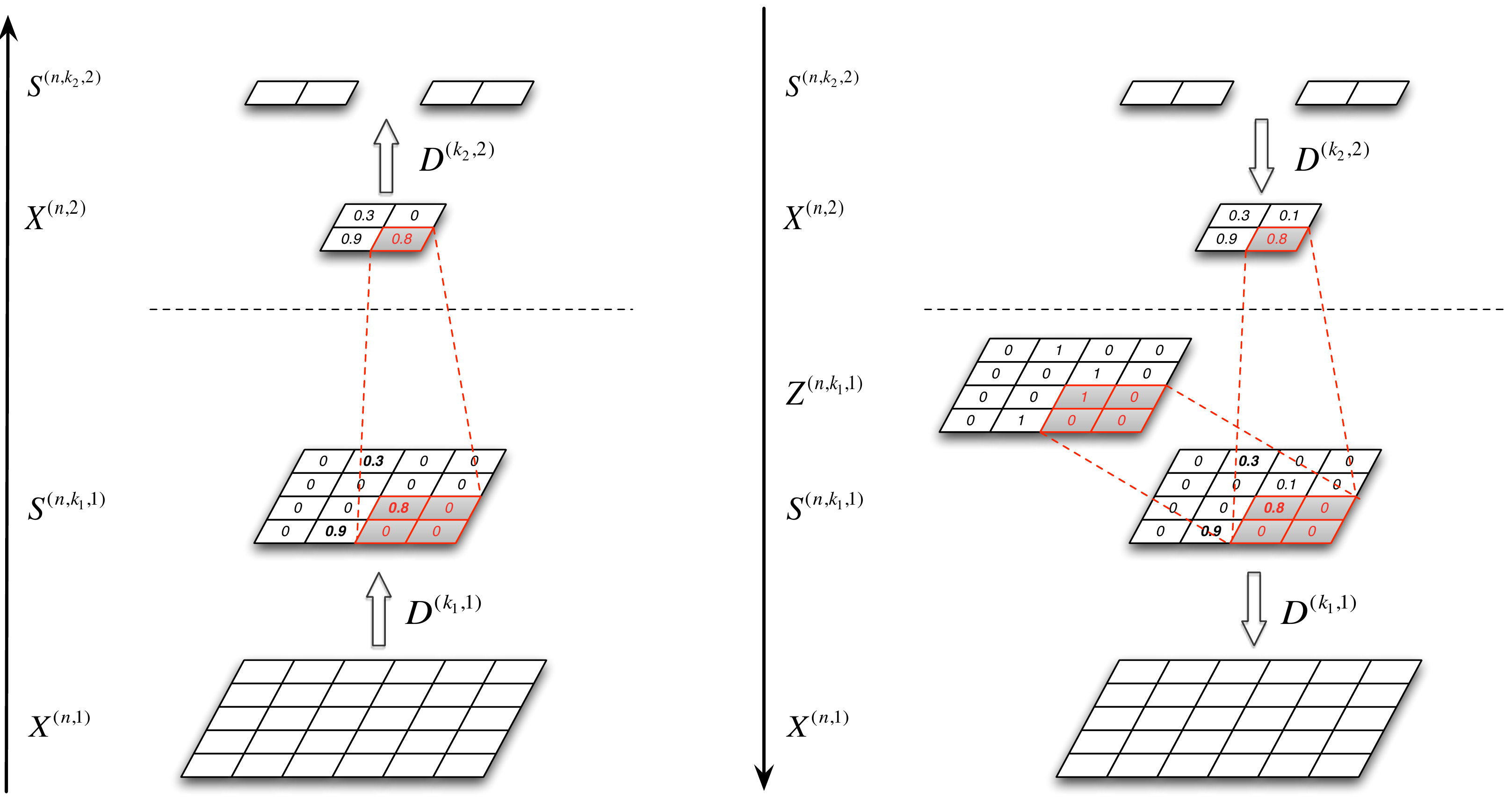}
	\vspace{-3mm}
	\caption{\small{Schematic of the proposed generative process. Left: bottom-up pretraining, right: top-down refinement. (Zoom-in for best visulization and a larger version can be found in the Supplementary Material.)}}
	\vspace{-6mm}
	\label{fig:max_pool}
\end{figure}

We now discuss the relationship between ${\bf S}^{(n,k_l,l)}$ and layer $k_l$ of ${\bf X}^{(n,l+1)}$. The 2D activation map ${\bf S}^{(n,k_l,l)}$ is partitioned into $n_x\times n_y$ dimensional contiguous blocks (pooling blocks with respect to layer $l+1$ of the model); see the left part of Figure \ref{fig:max_pool}. Associated with each block of pixels in ${\bf S}^{(n,k_l,l)}$ is one pixel at layer $k_l$ of ${\bf X}^{(n,l+1)}$; the relative locations of the pixels in ${\bf X}^{(n,l+1)}$ are the same as the relative locations of the blocks in ${\bf S}^{(n,k_l,l)}$. Within each block of ${\bf S}^{(n,k_l,l)}$, either all $n_xn_y$ pixels are zero, or only one pixel is non-zero, with the position of that pixel selected stochastically via a multinomial distribution. Each pixel at layer $k_l$ of ${\bf X}^{(n,l+1)}$ equals the largest-amplitude element in the associated block of ${\bf S}^{(n,k_l,l)}$ ($i.e.$, max pooling). Hence, if all elements of a block of ${\bf S}^{(n,k_l,l)}$ are zero, the corresponding pixel in ${\bf X}^{(n,l+1)}$ is also zero. If a block of ${\bf S}^{(n,k_l,l)}$ has a (single) non-zero element, that non-zero element is the corresponding pixel value at the $k_l$-th layer of ${\bf X}^{(n,l+1)}$.

The bottom-up generative process for each block of ${\bf S}^{(n,k_l,l)}$ proceeds as follows (left part of Figure \ref{fig:max_pool}). The model first imposes that a given block of ${\bf S}^{(n,k_l,l)}$ is either all zero or has one non-zero element, and this binary question is modeled as the beta-Bernoulli representation of (\ref{Eq:x_l}). If a given block has a non-zero value, the position of that value in the associated $n_x\times n_y$ block is defined by a multinomial distribution, and its value is modeled as $w_{i,j}^{(n,k_l,l)}$ represented in (\ref{Eq:x_l}). The beta-Bernoulli step, followed by multinomial, are combined into one equivalent statistical representation, as discussed next.

%

Let ${\bf z}^{(n,k_l,l)}_{i^{\prime}, j^{\prime}}\in \{0,1\}^{n_xn_y}$ denote the $(i^{\prime}, j^{\prime})$-th block of ${\bf Z}^{(n,k_l,l)}$ at layer $l$, where $i^{\prime} = 1, \dots, \frac{N_x}{n_x}; j^{\prime} = 1,\dots, \frac{N_y}{n_y}$ assuming integer divisions. 
We introduce a latent variable ${\bf c}^{(n,k_l,l)}_{i^{\prime}, j^{\prime}}\in\{0,1\}^{n_xn_y +1}$ to implement at most one non-zero element out of the $n_xn_y$ entries in $\{{ z}^{(n,k_l,l)}_{i^{\prime}, j^{\prime},m}\}_{m=1}^{n_xn_y}$ through 
\begin{equation}
{z}^{(n,k_l,l)}_{i^{\prime}, j^{\prime},m} = { c}^{(n,k_l,l)}_{i^{\prime}, j^{\prime},m}, \quad\quad
{\bf c}^{(n,k_l,l)}_{i^{\prime}, j^{\prime}} \sim {\rm Mult}(1; \thetav^{(n,k_l,l)}), \quad\quad
  {\thetav}^{(n,k_l,l)} \sim {\rm Dir}\left(\frac{1}{n_x n_y +1}\right),
\end{equation}
where ${\rm Mult}(\cdot)$ and ${\rm Dir}(\cdot)$ denote multinomial and Dirichlet distribution, respectively (the Dirichlet distribution has a {\em set} of parameters, and here we imply that are equal, and set to the value indicated in ${\rm Dir}(\cdot)$).
${\bf c}^{(n,k_l,l)}_{i^{\prime}, j^{\prime}}$ has $(n_xn_y +1)$ entries, of which only one is equal to 1. If the last element is 1, this means all $\{{ z}^{(n,k_l,l)}_{i^{\prime}, j^{\prime}, m}\}_{m=1}^{n_x n_y} = 0$.
Since the $(i^{\prime}, j^{\prime})$-th block at layer $l$ corresponds to one element at layer $(l+1)$,
we have 
\begin{equation} \label{Eq:Sxz}
s^{(n,k_{l}, l)}_{i^{\prime},j^{\prime},m}=  {x}^{(n,k_l, l+1)}_{i^{\prime}, j^{\prime}}  z^{(n,k_{l},l)}_{i^{\prime}, j^{\prime},m}, ~~\forall m=1,\dots, n_xn_y 
\end{equation}
Hence, if the last element of ${\bf c}^{(n,k_l,l)}_{i^{\prime}, j^{\prime}}$ is 1, all elements of block $(i^\prime,j^\prime)$ are zero; if not, the location of the non-zero element in the first $n_xn_y$ entries of ${\bf c}^{(n,k_l,l)}_{i^{\prime}, j^{\prime}}$ locates the position of the non-zero element in the corresponding block. The remaining parts of the model are represented as in (\ref{Eq:x_l}).

In the pretraining phase, we start with ${\bf X}^{(n,1)}$, which is the data ${\bf X}^{(n)}$. We learn $\{ {\bf S}^{(n,k_1,1)}\}_{k_1=1,K_1}$ using the blocked activation weights, via Gibbs sampling, where the multinomial distribution associates each non-zero element with a position in the corresponding block. The MAP Gibbs sample is then selected, defining model parameters for the layer under analysis. The ``stacked'' and pooled $\{{\bf S}^{(n,k_1,1)}\}_{k_1=1,K_1}$ are used to define ${\bf X}^{(n,2)}$, and the learning procedure then continues, learning dictionary elements ${\bf D}^{(k_2,2)}$ and activation maps $\{{\bf S}^{(n,k_2,2)}\}_{k_2=1,K_2}$, again via Gibbs sampling and MAP selection. This continues sequentially up to the $L$-th, or top, layer.
For the top layer, since no pooling is necessary, the beta-Bernoulli prior in (\ref{eq:beta-bern}) is used.

\vspace{-2mm}
\subsection{Model Refinement With Stochastic Pooling\label{sec:refine}}
\vspace{-3mm}

The learning performed with the top-down generative model (right part of Fig.~\ref{fig:max_pool}) constitutes a {\em refinement} of the parameters learned during pretraining, and the excellent initialization constituted by the parameters learned during pretraining is key to the subsequent model performance.


In the refinement phase, the equations are (almost) the same, but we now proceed top down, from (\ref{Eq:x_lp1}) to (\ref{Eq:x_l}). The generative process constitutes ${\bf D}^{(k_{l+1}, l+1)}$ and ${\bf Z}^{(n,k_{l+1},l+1)}  \odot {\bf W}^{(n,k_{l+1}, l+1)}$, and after convolution ${\bf X}^{(n,l+1)}$ is manifested; the ${\bf E}^{(n,l)}$ is now absent at all layers, except layer $l=1$, at which the fit to the data is performed. Each element of ${\bf X}^{(n,l+1)}$ has an associated pooling {\em block} in ${\bf S}^{(n,k_l,l)}$. Via a multinomial distribution like in pretraining, each element of ${\bf X}^{(n,l+1)}$ is mapped to one position in the corresponding block of ${\bf S}^{(n,k_l,l)}$, and all other elements in that $n_x\times n_y$ block are set to zero. Since ${\bf X}^{(n,l+1)}$ is manifested top-down as a convolution of ${\bf D}^{(k_{l+1}, l+1)}$ and ${\bf Z}^{(n,k_{l+1},l+1)}  \odot {\bf W}^{(n,k_{l+1}, l+1)}$, ${\bf X}^{(n,l+1)}$ will in general have no elements exactly equal to zero (but many will be small, based on the pretraining). Hence, {\em each} block of ${\bf S}^{(n,k_l,l)}$ will have one non-zero element, with position defined by the multinomial\footnote{We also considered a model exactly like in pretraining, which in the pooling step a pixel in ${\bf X}^{(n,l+1)}$ could be mapped via the multinomial to an all-zero activation block in layer $l$; the results are essentially unchanged from the method discussed above.}.

During pretraining many blocks of ${\bf S}^{(n,k_l,l)}$ will be all-zero since we preferred a sparse representation, while during refinement this sparsity requirement is relaxed, and in general each pooling block of ${\bf S}^{(n,k_l,l)}$ will have one non-zero element (but it is still sparse), and this value is mapped via pooling to the corresponding pixel in ${\bf X}^{(n,l)}$. In pretraining the Dirichlet and multinomial distributions were of size $n_xn_y+1$, allowing the all-zero activation block; during refinement the multinomial and Dirichlet are of dimensions $n_xn_y$. The corresponding $n_xn_y$ Dirichlet and multinomial parameters from pretraining are used to constitute initializations for refinement.

\vspace{-4mm}
\subsection{Top-Level Features and Testing}
\vspace{-3mm}
In order to understand deep convolutional models, researchers have visualized dictionary elements mapped to the image level~\citep{Zeiler14ECCV}. One key challenge of this visualization is that one dictionary element at high layers can have {\em multiple} representations at the layer below, given different activations in each pooling block (in our model, this is manifested by the stochasticity associated with the multinomial-based pooling). \cite{Zeiler14ECCV} showed different versions of the same upper-layer dictionary element at the image level. Because of this capability of accurate dictionary localization at each layer, deep convolutional models perform well in classification.
However, also due to these multiple representations, during testing, one has to infer dictionary activations layer by layer (via deconvolution), which is computationally expensive.
In order to alleviate this issue, \cite{LeCun10NIPS} proposed an approximation method using convolutional filter banks (fast because there is no explicit deconvolution) followed by a nonlinear function. 
Though efficient at test time, in the training step one must simultaneously learn deconvolutional dictionaries and associated filterbanks, and the choice of non-linear function is critical to the performance of the model. Moreover, in the context of the framework proposed here, it is difficult to integrate the approach of \cite{Kavukcuoglu08,LeCun10NIPS} into a Bayesian model. 

We propose a new approach to accelerate testing. 
After performing model learning (after refinement), we project top-layer dictionary elements down to the data plane. At test, deconvolution is only performed once, using the top-layer dictionary elements mapped to the data plane. The top-layer activation strengths inferred via this deconvolution are then used in a subsequent classifier. The different manifestations of a top-layer dictionary element mapped to the data plane are constituted by different (stochastic) pooling mappings via the multinomial. To select top-layer dictionary elements in the data plane, used for test, we employ maximum-likelihood (ML) dictionary elements, with ML performed across the different choices of the max pooling at each layer. Hence, after this ML-based top-layer dictionary selection, a pixel at layer $l+1$ is mapped to the same location in the associated layer $l$ block, for all convolutional shifts (same max-pooling map for all shifts at a given layer). Hence, the key approximation is that the stochastic pooling employed for each pixel at layer $l+1$ to a position in a block at layer $l$ is replaced by an ML-based \emph{deterministic} pooling (possibly a different deterministic map at each layer).
This simple approach has the advantage of \cite{Zeiler14ECCV} at test, in that we retain the deconvolution operation (unlike \cite{LeCun10NIPS}), but deconvolution must only be performed \emph{once} (not at each layer). In the experiments presented below, when visualizing inferred dictionary elements in the image plane, this ML-based dictionary selection is employed.  More details on this aspect of the model are provided in the Supplementary Material.

\vspace{-5mm}
\section{Gibbs-Sampling-Based Learning and Inference}
\vspace{-4mm}
Due to  local conjugacy at every component of the model, the local conditional posterior distribution for all parameters of our model is manifested in closed form, yielding efficient Gibbs sampling (see Supplementary Material for details). As in all previous convolutional models of this type, the FFT is leveraged to accelerate computation of the convolution operations, here within Gibbs update equations.

In the pre-training step, we select the ML sample from 500 collection samples, after first computing and discarding 1500 burn-in samples.
The same number of burn-in and collection samples, with ML selection, is performed for model refinement.
This ML selection of collection samples shares the same spirit as \cite{Geman1984}, in the sense of yielding a MAP solution (\emph{not} attempting to approximate the full posterior). 
During testing, we select the ML sample across 200 deconvolutional samples, after first discarding 500 burn-in samples.

\vspace{-4mm}
\section{Experimental Results}
\label{Sec:Exp}
\vspace{-4mm}
We here apply our model to the MNIST and Caltech 101 datasets.
We compare dictionaries (viewed in the data plane) before and after refinement.
Classification results (average of 10 trials) using top-layer features are presented for both datasets.
As in \citep{Paisley09ICML}, the hyperparameters are set as $a_0=1/K, b_0=1-1/K$, where $K$ is the number of dictionary elements at the corresponding layer, and $a_w=b_w=a_d=b_d=a_e=b_e=10^{-6}$; these are standard hyperparameter settings \citep{Paisley09ICML} for such models, and no tuning or optimization was performed.
All code is written in MATLAB and executed on a desktop with 3.8 GHz CPU and 24G memory.
Model training including refinement with one class (30 images) of Caltech 101 takes about 40 CPU minutes, and
testing (deconvolution) for one image takes less than 1 second. These results were run on a single computer, for demonstration, and acceleration via parallel implementation, GPUs~\citep{HintonNIPS2012}, and coding in C will be considered in the future; the successes realized recently in accelerating convolution-based models of this type are transferrable to our model.

\vspace{-3mm}
\paragraph{MNIST Dataset}
\begin{wraptable}{r}{0.55\textwidth}
	\vspace{-6mm}
	\caption{\small{Classification Error of MNIST data}}
	\vspace{-4mm}
	\centering
		\small
	\begin{tabular}{c|c}
		Methods & Test error \\
		\hline
		DBN~\cite{Hinton06Science} & 1.20\% \\
		\hline
		CBDN~\cite{Lee09ICML} & 0.82\%\\
		\hline
			$\begin{array}{l}
			\text{2-layer Conv. Net + 2-layer}\\
			\text{Classifier~\cite{Jarrett09ICCV}} \end{array}$&  0.53\% \\
		\hline 
		$\begin{array}{l}
		\text{6-layer Conv. Net + 2-layer Classifier } \\
		\text{+ elastic distortions~\cite{Ciresan11IJCAI}}
		\end{array}$
		& 0.35\% \\
		\hline
			MCDNN~\cite{ciresan2012multi} & 0.23\%\\
		\hline
		SPCNN~\cite{Zeiler13ICLR} & \\ 
		Average Pooling & 0.83\% \\
		Max Pooling & 0.55\% \\
		Stochastic Pooling & 0.47\%\\
		\hline
		$\begin{array}{l}
		\text{HBP~\cite{Chen13deepCFA},}\\
		\text{2-layer cFA + 2-layer features} \end{array}$& \\
		MCMC (10000 Training) & 0.89\% \\
		Batch VB (10000 Training) & 0.95\%  \\
		online VB (60000 Training) & 0.96\% \\
		\hline
		Ours, 2-layer model + 1-layer features & \\
		60000 Training & 0.42\%  \\
		10000 Training & 0.68\%  \\
		5000 Training & 1.02\% \\
		2000 Training & 1.11\% \\
		1000 Training & 1.66\%
	\end{tabular}
	\label{Table:Error_MNIST}
	\vspace{-10mm}
\end{wraptable}
%
We first consider the widely studied MNIST data (\url{http://yann.lecun.com/exdb/
mnist/}), which has 60,000 training and 10,000
testing images, each $28\times28$, for digits 0 through 9.
A two layer model is used with dictionary size $8\times 8$ ($n_{d_x}=n_{d_y}=8$) at the first layer and $6\times 6$ at the second layer; the pooling size is $3\times 3$ ($n_x=n_y=3$) and the number of dictionary elements at layers 1 and 2 are $K_1=32$ and $K_2=160$, respectively. 
We obtained these number of dictionary elements via setting the initial dictionary number to a relatively large value in the pre-training step and discarding infrequently used elements by counting the corresponding binary indicator ${\bf Z}$, {\em i.e.}, \emph{inferring} the number of needed dictionary elements, as in~\cite{Chen13deepCFA}. 

\begin{figure}[htbp!]
\centering
\vspace{-4mm}
\setcounter{subfigure}{0}
\subfloat[]{\label{Fig:MNIST_dict}\includegraphics[width=\textwidth, height=2.5cm]{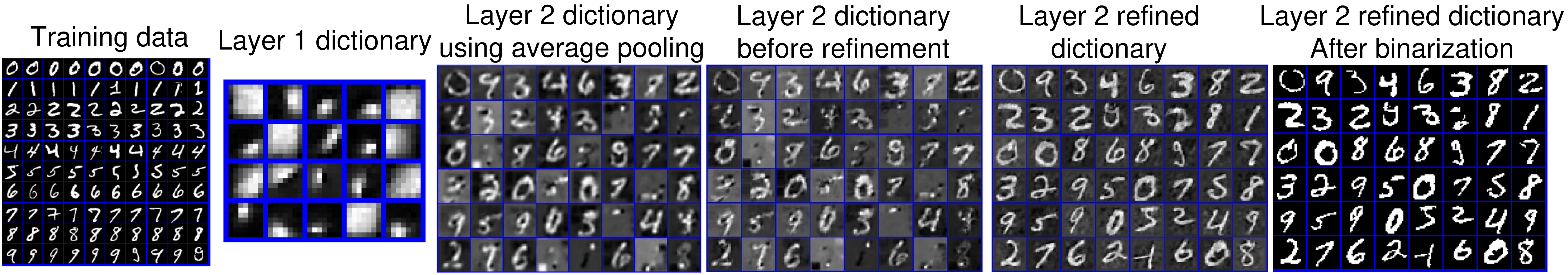}}\\
\vspace{-4mm}
\subfloat[]{\label{Fig:MNIST_missing}\includegraphics[width=0.65\textwidth, height=5.3cm]{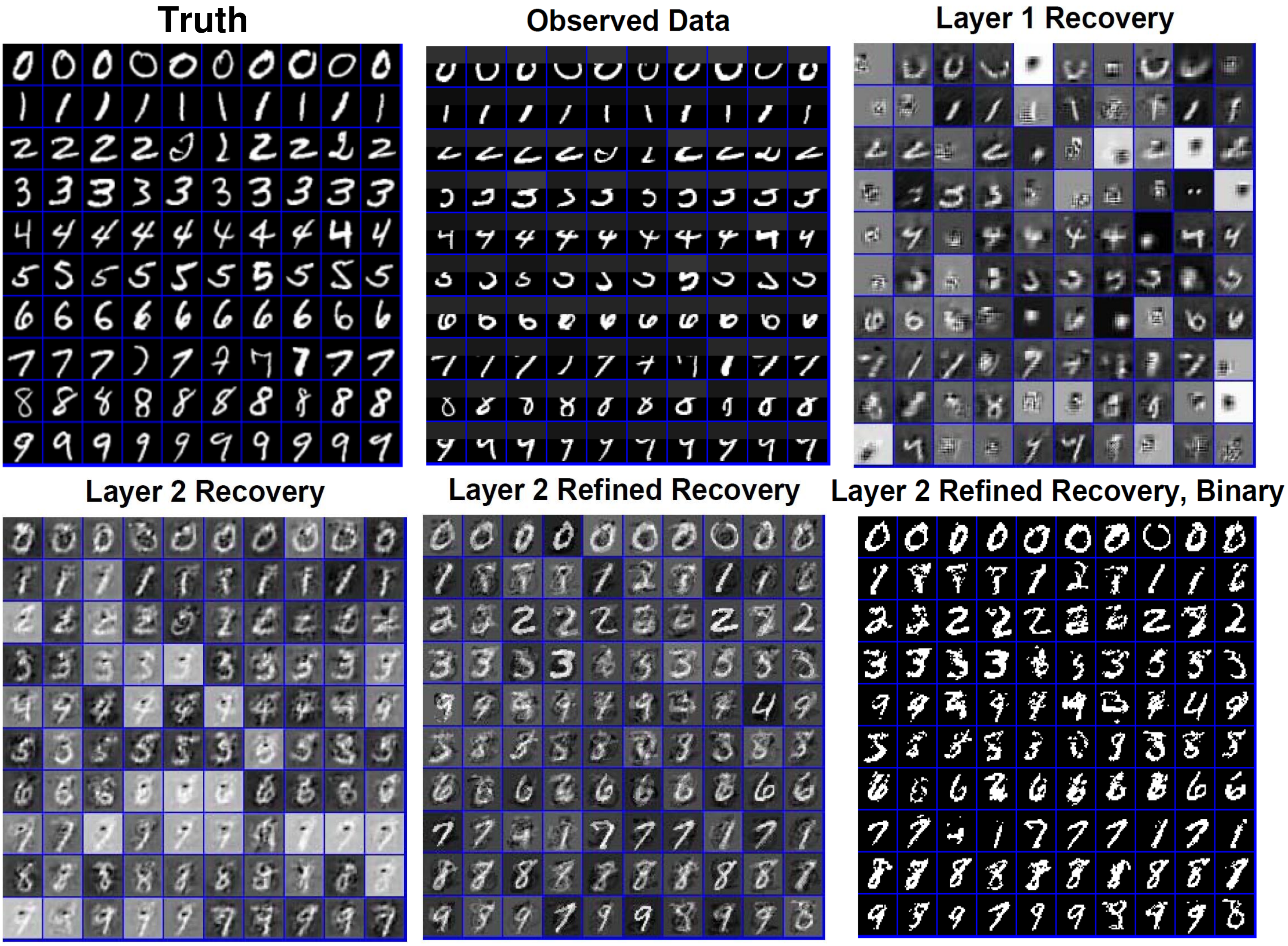}}
\quad~~
\subfloat[]{\label{Fig:MNIST_missing2}\includegraphics[width=0.3\textwidth,height=5.3cm]{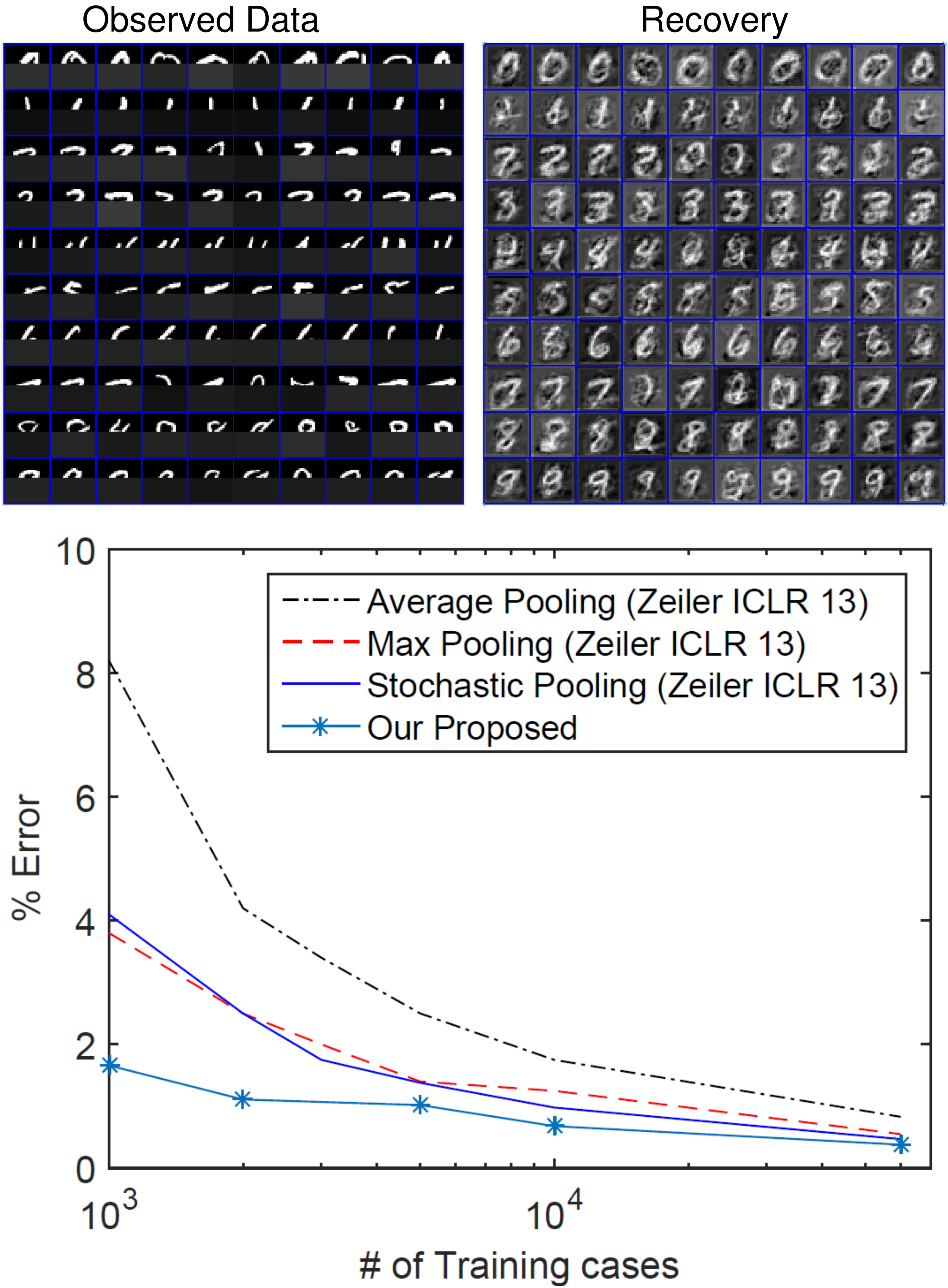}}
\vspace{-4mm}
\caption{\small{(a) Visualization of the dictionary learned by the proposed model. Note the refined dictionary (right) is much sharper than the dictionary before refinement (middle). (b) Missing data interpolation of digits.
(c) Upper part: a more challenging case for missing data interpolation of digits. Bottom part: testing error when training with reduced dataset sizes on MNIST.}
}
\vspace{-6mm}
\label{fig:reconSpec}
\end{figure}

Table~\ref{Table:Error_MNIST} summaries the classification results of our model compared with some related results, on the MNIST data.
The second (top) layer features corresponding to the refined dictionary are sent to a nonlinear
support vector machine (SVM)~\citep{CC01a} with Gaussian kernel, in a one-vs-all multi-class classifier, with classifier parameters tuned via 5-fold cross-validation (no tuning on the deep feature learning).
Rather than concatenating features at all layers as in~\cite{Zeiler13ICLR,Chen13deepCFA}, we only use the top layer features as the input to the SVM (deconvolution is only performed with top-layer dictionary elements), which saves much computation time (as well as memory) in both inference and classification, since the feature size is small.  
When the model is trained using all 60000 digits, we achieve an error rate of $0.42\%$ on testing, which is very close to the state-of-the-art, but with a relatively simpler model compared to~\cite{ciresan2012multi}; the error rate obtained using features learned after pretraining, before refinement, are similar to those in \cite{Chen13deepCFA} ($0.9\%$ error), underscoring the importance of the refinement step. 
We further plot the testing error in 
Fig.~\ref{Fig:MNIST_missing2} (bottom part) when the training size is reduced compared to the results reported in~\cite{Zeiler13ICLR}.
It can be seen that our model outperforms every approach in~\cite{Zeiler13ICLR}.
%
In order to examine the properties of the learned model, in Fig.~\ref{Fig:MNIST_dict} we visualize trained dictionaries at layer 2 mapped down to the data level. 
It is observed qualitatively that refinement improves the dictionary; the atoms after refinement are much sharper. 
If the average pooling described in~\cite{Zeiler13ICLR} is used, the dictionaries are blurry (middle-left part of Fig.~\ref{Fig:MNIST_dict}). 
When a threshold is imposed on the refined dictionary elements, they look like digits (rightmost part).  

To further verify the efficacy of our model, we show in Fig.~\ref{Fig:MNIST_missing} the interpolation results of digits with half missing, as in~\cite{Lee09ICML}. A one-layer model cannot recover the digits, while a two-layer model provides a good recovery (bottom row of Fig.~\ref{Fig:MNIST_missing}). Furthermore, by using our refinement approach, the recovery is much clearer (comparing the bottom-left part and bottom-middle part of Fig.~\ref{Fig:MNIST_missing}).
Given this excellent performance, more challenging interpolation results are shown in Fig.~\ref{Fig:MNIST_missing2} (upper part), where we cannot identify any digits from the observations; even in this case, the model can provide promising reconstructions.

\begin{figure}[htbp!]
  \centering
  \vspace{-3mm}
  \includegraphics[width=\textwidth]{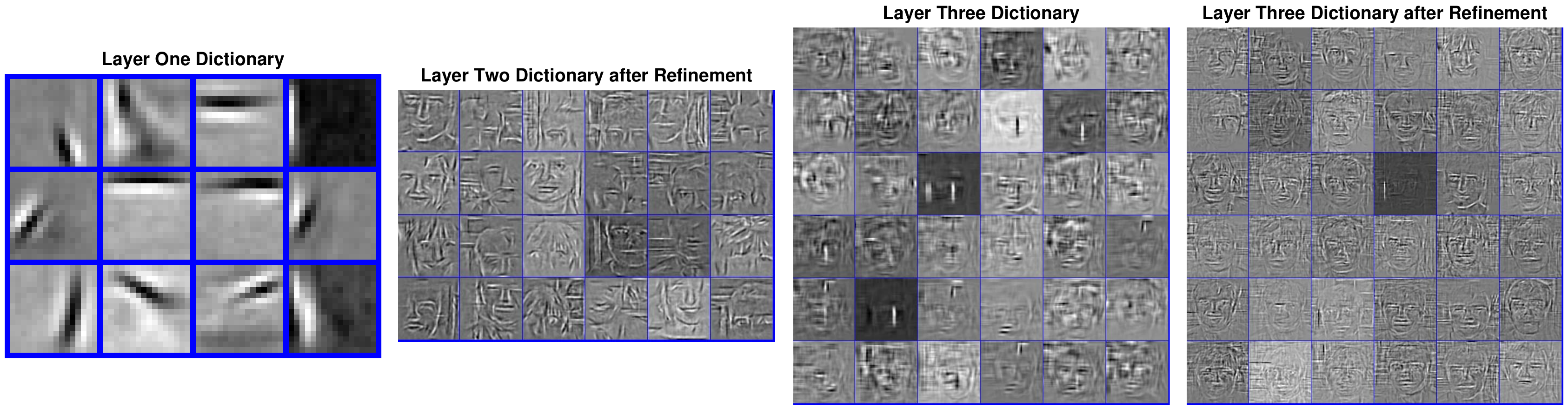}
  \vspace{-6mm}
  \caption{\small{Dictionary elements in each layer trained with 64 ``face easy" images from Caltech 101}.}
    \label{Fig:face_train}
    \vspace{-4mm}
\end{figure}
\begin{figure}[htbp!]
  \centering
  \vspace{-1mm}
  \includegraphics[width=\textwidth]{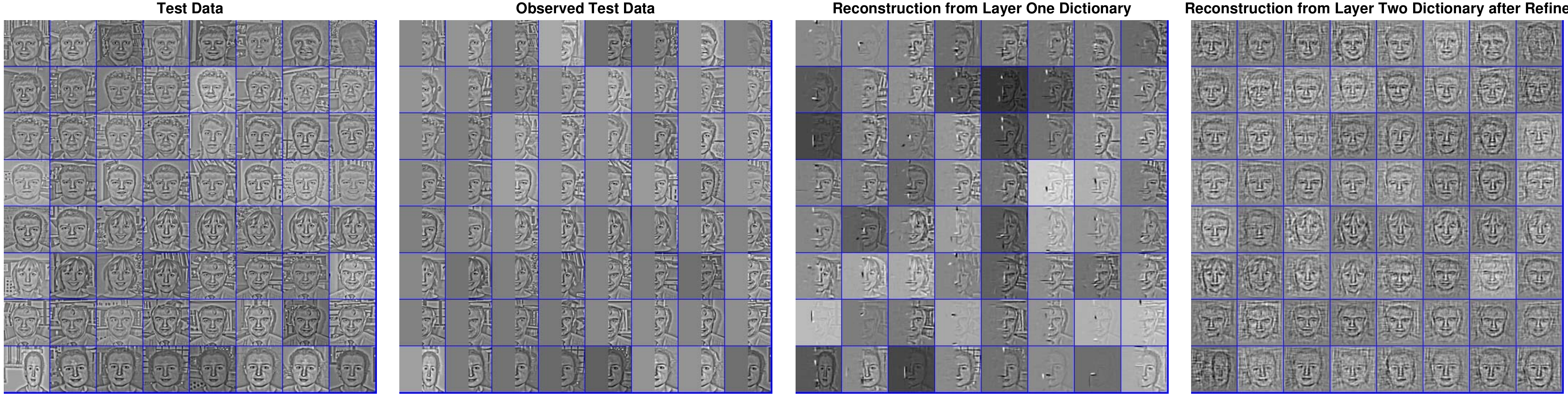}
   \vspace{-7mm}
  \caption{\small{Face data interpolation using a 2-layer model. From left to right: truth, observed data, layer-1 recovery, layer-2 recovery.} }
    \label{Fig:face_miss}
    \vspace{-3mm}
\end{figure}

%

\vspace{-3mm}
\paragraph{Caltech 101 Dataset}
We next consider the Caltech 101 dataset.
First we analyze our model with images in the ``easy face" category;  
64 images (after local contrast normalization~\citep{Jarrett09ICCV}) have been resized to $128\times 128$ and a three-layer deep model is used.
At layers 1, 2 and 3, the number of dictionary elements is set respectively to $K_1=16$, $K_2=24$ and $K_3=36$ (these inferred in the pretraining step, as discussed above), with dictionary sizes $17\times 17$, $9\times 9$ and $6\times 6$. 
The pooling sizes are $4\times 4$ (layer 1 to layer 2) and $2\times 2$ (layer 2 to layer 3). 
Example learned dictionary elements are mapped to the image level and shown in Fig.~\ref{Fig:face_train}. It can be seen that the first-layer dictionary extracts edges of the images,
while the second-layer dictionary elements look like a part of the face and the third-layer elements are almost entire faces.
We can see the improvement manifested by refinement by comparing the right two parts in Fig.~\ref{Fig:face_train} (the dictionaries after refinement are sharper).
Similar to the MNIST example, we also show in Fig.~\ref{Fig:face_miss} the interpolation results of face data with half missing, using a two-layer model (the dictionary sizes are $14\times 14$ and $13\times 13$ at layers 1 and 2, respectively, with max-pooling size $3\times 3$.).
It can be seen the missing parts are recovered progressively more accurately considering a one- and two-layer model.
Though the background is a little noisy, each face is recovered in great detail by the second layer dictionary (a three-layer model gives similar results, omitted here for brevity).
%
\begin{figure}[htbp!]
	\centering
	\vspace{-3mm}
	\includegraphics[width=\textwidth,height = 5cm]{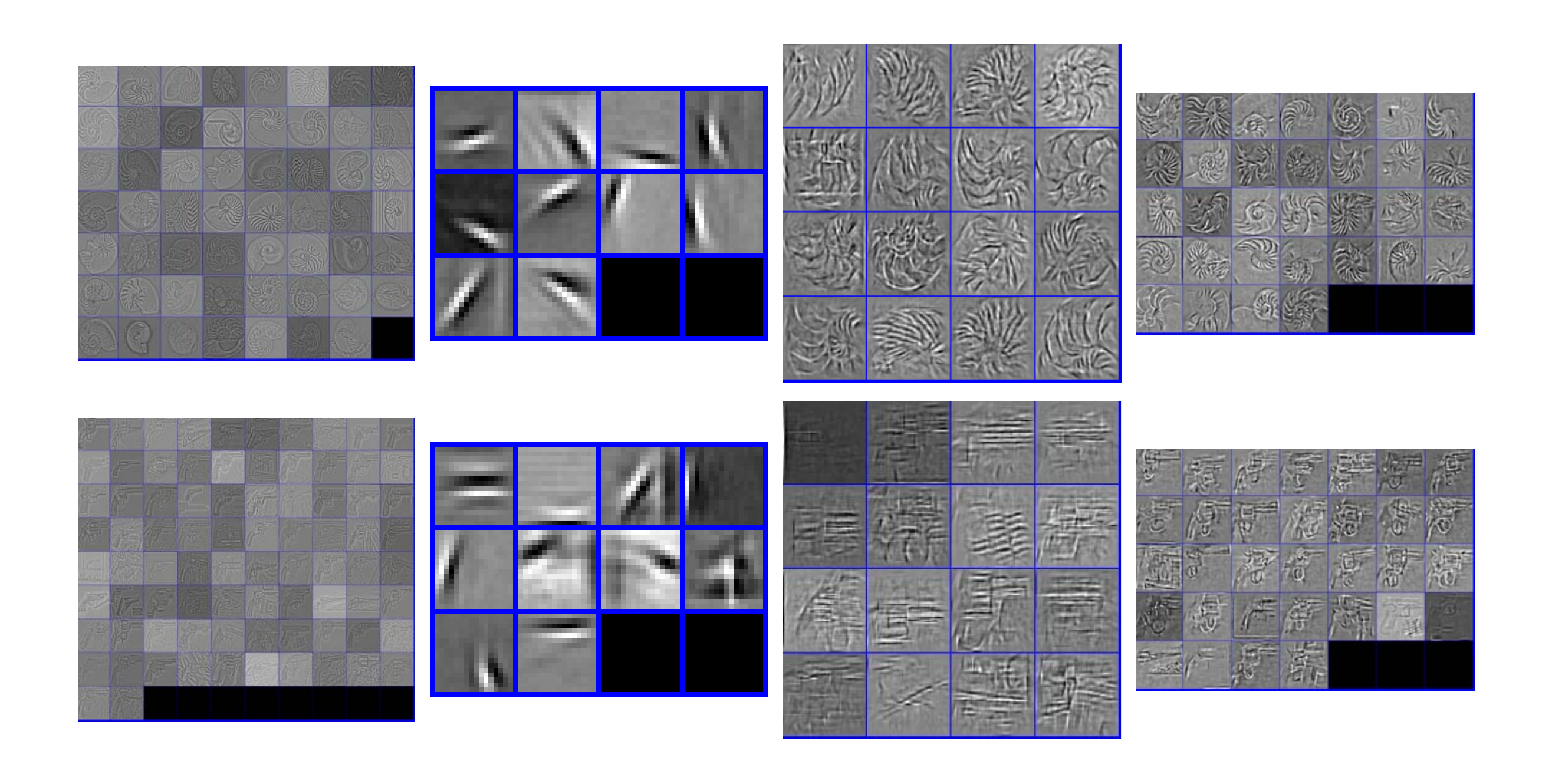}
	\vspace{-5mm}
	\caption{\small{Trained dictionaries per class mapped to the data plane. Row 1-2: nautilus, revolver. Column 1-4: training images after local contrast normalization, layer-1 dictionary, layer-2 dictionary, layer-3 dictionary.}}
	\label{Fig:SepDict}
	\vspace{-5mm}
\end{figure}
%

We develop Caltech 101 dictionaries by learning on each data class in isolation, and then concatenate all (top-layer) dictionaries when learning the classifier. 
In Figure~\ref{Fig:SepDict} we depict dictionary elements learned for two data classes, projected to the image level (more results are shown in the Supplementary Material). It can be seen the layer-1 dictionary elements are similar for the two data classes, while the upper-layer dictionary elements are data-class dependent.
One problem of this parallel training is that the dictionary may be redundant across image classes (especially at the first layer). However, during testing, using the proposed approach, we only use top-layer dictionaries, which are typically distinct across data classes (for the data considered).

\begin{wraptable}{r}{0.56\textwidth}
\vspace{-4mm}
	\caption{ \small{Classification Accuracy Rate of Caltech-101.}}
	\vspace{-0.3cm}
	\centering
	\small
	\begin{tabular}{c|c|c}
		\# Training Images per Category & 15 & 30 \\
		\hline
		DN~\cite{Zeiler10CVPR}  & 58.6 \% & 66.9\% \\
		\hline
		CBDN~\cite{Lee09ICML} & 57.7 \% & 65.4\% \\
		\hline
		HBP ~\cite{Chen13deepCFA}  & 58\% & 65.7\% \\
		\hline
		ScSPM ~\cite{yang09CVPR} & 67 \% & 73.2\%  \\
		\hline
		P-FV ~\cite{seidenari2014local} & 71.47\% & 80.13\% \\
		\hline
		R-KSVD ~\cite{li2013reference}  & 79 \% & 83\% \\
		\hline
		Convnet~\cite{Zeiler14ECCV}  & 83.8 \% & 86.5\% \\	
		\hline
		Ours, 2-layer model + 1-layer features  & 70.02\% & 80.31\%  \\
		\hline
		Ours, 3-layer model + 1-layer features & 75.24\% & 82.78\% 
	\end{tabular}
	\label{Table:accuracy_caltech101}
	\vspace{-0.4cm}
\end{wraptable}
For Caltech 101 classification, we follow the setup in~\cite{yang09CVPR}, selecting 15 and 30 images per category for training, and testing on the rest.
The features of testing images are inferred based on the top-layer dictionaries and sent to a multi-class SVM; we again use a Gaussian kernel non-linear SVM with parameters tuned via cross-validation.
Ours and related results are summarized in Table~\ref{Table:accuracy_caltech101}.
For our model, we present results based on 2-layer and 3-layer models.
It can be seen that our model (the 3-layer one) provides results close to the state-of-the-art in~\cite{Zeiler14ECCV}, which used a much more complicated model ({\em i.e.}, a 7-layer convolutional network and used the ImageNet dataset to pretrain the network), and our results are also very close to the state-of-the-art results using hand-crafted features ({\em e.g.}, SIFT in \cite{li2013reference}).
Based on features learned by our model at the pretraining stage, our classification performance is comparable to that of the HBP model in \cite{Chen13deepCFA} (around 65\% accuracy for a 2-layer model, when training with 30 examples per class), with our results demonstrating a 17\% improvement in performance after model refinement.

\vspace{-3mm}
\section{Conclusions}
\vspace{-4mm}
A deep generative convolutional dictionary-learning model has been developed within a Bayesian setting, with efficient Gibbs-sampling-based MAP parameter estimation. The proposed framework enjoys efficient bottom-up and top-down probabilistic inference. A probabilistic pooling module has been integrated into the model, a key component to developing a principled top-down generative model, with efficient learning and inference.
Extensive experimental results demonstrate the efficacy of the model to learn multi-layered features from images. 
A novel method has been developed to project the high-layer dictionary elements to the image level, and efficient single-layer deconvolutional inference is accomplished during testing. On the MNIST and Caltech 101 datasets, our results are very near the state of the art, but with relatively simple model complexity at test.
Future work includes performing deep feature learning and classifier design jointly. The algorithm will also be ported to a GPU-based implementation, allowing scaling to large-scale datasets.

\bibliography{iclr2015}

\newpage
\begin{center}
\huge{Supplementary Material}
\end{center}

\section{Conditional Posteriori Distributions for Gibbs Sampling \label{Sec:Gibbs}}
	In the $l-$th layer, the model can be formed as:	
\begin{align}
{\bf X}^{(n,k_{l-1}, l)} = \sum_{k_{l}=1}^{K_{l}} {\bf D}^{(k_{l}, l)} * \left({\bf Z}^{(n,k_{l},l)}  \odot {\bf W}^{(n,k_{l}, l)}\right) + {\bf E}^{(n, k_{l-1},l)} \label{Eq:x_lp2}
\end{align}
For simplification, we define the following symbols (operations):
\begin{align}
	X^{-(n,k_{l-1},l)} &= X^{(n,k_{l-1},l)} - \sum_{ k_l} D^{(k_{l-1},k_l,l)} * (Z^{(n,k_l,l)} \odot W^{(n,k_l,l)})\\
	X_{-k_l}^{(n,k_{l-1},l)} &= X^{(n,k_{l-1},l)} - \sum_{t \neq k_l} D^{(k_{l-1},t,l)} * (Z^{(n,t,l)} \odot W^{(n,t,l)})\\
	(D^{(k_{l-1},k_l,l)})^2&=D^{(k_{l-1},k_l,l)} \odot   D^{(k_{l-1},k_l,l)}\\
	(W^{(n,k_l,l)})^2&=W^{(n,k_l,l)}\odot   W^{(n,k_l,l)}\\
	(X^{-(n,k_{l-1},l)})^2&=X^{-(n,k_{l-1},l)}\odot    X^{-(n,k_{l-1},l)}
	\end{align}
	The symbol $\odot$ is the element-wise product operator and $\oslash$ is the element-wise division operator.
	
	$A=B \circledast C$ means if $B \in R^{n_B \times n_B}$ and $C \in R^{n_C \times n_C}$, $A \in R^{(n_B-n_C+1) \times (n_B-n_C+1)}$ with element $(i,j)$
	\begin{align}
	A_{i,j} = \sum_{p=1}^{n_C} \sum_{q=1}^{n_C} B_{p+i-1,q+j-1} C_{p,q}.
	\end{align}
For each MCMC iteration, the samples are drawn from:
\begin{itemize}
	\item
	Sample $D^{(k_{l-1},k_l,l)}$:
	\begin{align}
	D^{(k_{l-1},k_l,l)}|- &\sim N(\mu^{(k_{l-1},k_l,l)},\Sigma^{(k_{l-1},k_l,l)})\\
	\Sigma^{(k_{l-1},k_l,l)} &= 1 \oslash \left(\sum_{n=1}^{N} \gamma_e^{(n,k_{l-1},l)} \|Z^{(n,k_{l},l)} \odot W^{(n,k_{l},l)}\|_2^2 + \gamma_d^{(k_{l-1},k_l,l)}\right)\\
	\hat{\mu}^{(k_{l-1},k_l,l)} &= \Sigma^{(k_{l-1},k_l,l)} \odot \left[\sum_{n=1}^{N} \gamma_e^{(n,k_{l-1},l)} X_{-k_l}^{(n,k_{l-1},l)} \circledast (Z^{(n,k_{l},l)} \odot W^{(n,k_{l},l)})\right]
	\end{align}
	\item
	Sample $\gamma_d^{(k_{l-1},k_l,l)}$:
	\begin{align}
	\gamma_d^{(k_{l-1},k_l,l)}|- \sim \text{Gamma}\left(a_d+\frac{1}{2}, b_d+\frac{1}{2}(D^{(k_{l-1},k_l,l)})^2\right)
	\end{align}
	\item
	Sample $W^{(n,k_{l},l)}$:
	\begin{align}
	W^{(n,k_{l},l)}|- &\sim N(\xi^{(n,k_{l},l)}, \Lambda^{(n,k_{l},l)})\\
	\Lambda^{(n,k_{l},l)} &= 1 \oslash \left(\sum_{k_{l-1}=1}^{K_{l-1}} \gamma_e^{(n,k_{l-1},l)} \|D^{(k_{l-1},k_l,l)} \|_2^2 Z^{(n,k_{l},l)} + \gamma_W^{(n,k_{l},l)}\right)\\
	\xi^{(n,k_{l},l)} &= \Lambda^{(n,k_{l},l)} \odot Z^{(n,k_{l},l)} \odot \left[\sum_{k_{l-1}=1}^{K_{l-1}} \gamma_e^{(n,k_{l-1},l)} X_{-k_l}^{(n,k_{l-1},l)} \circledast D^{k_{l-1},k_l,l}\right]
	\end{align}
	\item
	Sample $\gamma_W^{(n,k_{l},l)}$
	\begin{align}
	\gamma_W^{(n,k_{l},l)}|- &\sim \text{Gamma}\left(a_W+\frac{1}{2}, b_W+\frac{1}{2}(W^{(n,k_l,l)})^2\right)
	\end{align}
	\item
	Sample $Z_{i',j'}^{(n,k_{l},l)}$:
	
	Let $m=1,...,n_xn_y$, $i=1,...,N_x$, $j=1,...,N_y$; $i'=1,...,N_x/n_x$, $j=1,...,N_y/n_x$; we can find $(i',j',m)$ and $(i,j)$ are one-to-one correspondence. From
	\begin{align}
	Y^{(n,k_{l},l)} = \sum_{k_{l-1}=1}^{K_{l-1}} \gamma_e^{(n,k_{l-1},l)} \left[ \|D^{(k_{l-1},k_{l},l)}\|_2^2 \odot \left(W^{(n,k_{l},l)}\right)^2 - 2 \left( X_{-k_l}^{(n,k_{l-1},l)} \circledast D^{k_{l-1},k_l,l} \right) \odot W^{(n,k_{l},l)} \right],
	\end{align}
	and
	\begin{align}
	\hat{\theta}_{i',j',m}^{(n,k_{l},l)} &= \theta_m^{(n,k_{l},l)} Y_{i',j',m}^{(n,k_{l},l)} \qquad \text{for } m=1,..., n_xn_y,
	\end{align}
	we have
	\begin{align}
	P(Z_{i',j',m}^{(n,k_{l},l)} = 1) &= \frac{\hat{\theta}_{i',j',m}^{(n,k_{l},l)}}{\sum_{t=1(\neq m)}^{n_xn_y}\hat{\theta}_{i',j',m}^{(n,k_{l},l)}+\theta_{n_xn_y+1}^{(n,k_{l},l)}}.
	\end{align}
	\item
	Sample $\theta^{(n,k_{l},l)}$:
	\begin{align}
	\theta^{(n,k_{l},l)}|- &\sim \text{Dirichlet} ({\alpha}^{(n,k_{l},l)})\\
	\alpha_m^{(n,k_{l},l)} &= \frac{1}{n_xn_y+1} + \sum_{i'} \sum_{j'} z_{i',j',m}^{(n,k_{l},l)} \qquad \text{for }~~~~ m=1,...,n_xn_y,\\
	\alpha_{n_xn_y+1}^{(n,k_{l},l)} &= \frac{1}{n_xn_y+1} + \sum_{i'} \sum_{j'} (1-\sum_m z_{i',j',m}^{(n,k_{l},l)})
	\end{align}
	\item
	Sample $\gamma_e^{(n,k_{l-1},l)}$:
	\begin{align}
	\gamma_e^{(n,k_{l-1},l)}|- &\sim \text{Gamma}\left(a_e+\frac{1}{2}, b_e+(X^{-(n,k_{l-1},l)})^2\right).
	\end{align}
\end{itemize}

\section{Projection of Dictionaries to the Data Layer \label{Sec:Proj}}
\subsection{Notation}
Assume $X\in {\mathbb R}^{N_x\times N_y}$ and $Y\in {\mathbb R}^{N_x/n_x\times N_y/n_y}$. Here $n_x, n_y\in N$ are the pooling ratio and the pooling map is $ Z\in \{0,1\}^{N_x\times N_y}$. In the $(i^{\prime}, j^{\prime})^{\rm th}$ block of $X$ and $ Z$, there is at most one non-zero element, where  $i^{\prime}\in\{1,...,\lceil N_x/n_x \rceil\}$, $j^{\prime}\in\{1,...,\lceil N_y/n_y\rceil\}$. Now, let $i\in\{1,...,N_x\}$, $j\in\{1,...,N_y\}$ then the following pooling and unpooling functions can be defined:
\begin{enumerate}[(1)]
	\item
	Define  $f:{\mathbb R}^{N_x\times N_y}\to {\mathbb R}^{N_x/n_x\times N_y/n_y}$, with $Y=f(X)$ . Recall that within each pooling block, $X$ has at most one non-zero element, and therefore
	\begin{align}
	Y_{i^{\prime},j^{\prime}}=\sum_{s=1}^{n_x}\sum_{t=1}^{n_y}X_{(i^{\prime}-1) n_x+s,(j^{\prime}-1) n_y+t}
	\end{align}
	The following is an example to demonstrate $f$:
	\begin{figure}[H]
		\centering
		\includegraphics[width=3in]{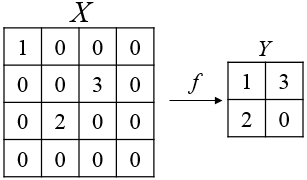}
	\end{figure}	
	\item
	Define  $f^{-1}:{\mathbb R}^{N_x/n_x\times N_y/n_y}\times \{0,1\}^{N_x\times N_y}\to {\mathbb R}^{N_x\times N_y}$, with $X=f^{-1}(Y,Z)$
	\begin{align}
	X_{i,j}=Y_{\left\lceil i/n_x\right\rceil, \left\lceil j/n_y\right\rceil}Z_{i,j}
	\end{align}
	The following is an example to demonstrate $f^{-1}$:
	\begin{figure}[H]
		\centering
		\includegraphics[width=4in]{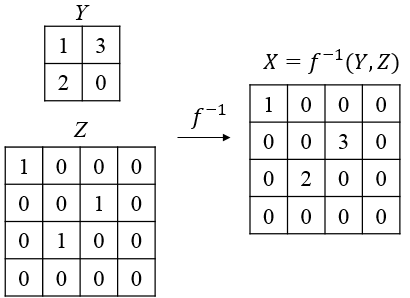}
	\end{figure}
	\item
	Define $g:{\mathbb R}^{N_x/n_x\times N_y/n_y}\times {\mathbb N}^2\to {\mathbb R}^{N_x\times N_y}$, with $A=g(Y,n_x, n_y)$
	\begin{align}
	A_{i,j}=Y_{\left\lceil \frac{i}{n_x}\right\rceil, \left\lceil \frac{j}{n_y}\right\rceil}.
	\end{align}
	The following is an example to demonstrate $g$,with $n_x=n_y=2$:
	\begin{figure}[H]
		\centering
		\includegraphics[width=3in]{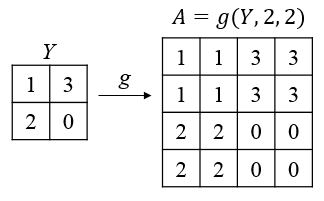}
	\end{figure}	
	\item
	Define $h:{\mathbb R}^{N_x/n_x\times N_y/n_y}\times {\mathbb N}^2\to {\mathbb R}^{((N_x-1)n_x+1)\times ((N_x-1)n_y+1)}$, with $B=h(Y,n_x,n_y)$:
	\begin{equation}
	B_{i,j} = \left\{
	\begin{array}{lcl}
	Y_{ \frac{i-1}{n_x}+1, \frac{j-1}{n_y}+1} &\text{if} &\frac{i-1}{n_x} \text{ and } \frac{j-1}{n_y} \text{ both are integers}\\
	0 &\text{otherwise} &
	\end{array}  
	\right.
	\end {equation}
	
	The following is an example to demonstrate $h$ with $n_x=n_y=2$:
	\begin{figure}[H]
	\centering
	\includegraphics[width=2.5in]{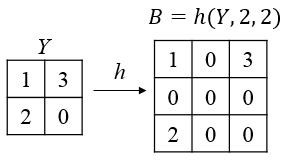}
	\end{figure}
\end{enumerate}
	\subsection{Some Useful Lemmas}
	\begin{lemma}
	$\displaystyle
	f^{-1}(Y,Z)=g(Y)\odot Z
	$
	\end{lemma}
	\begin{lemma}
	$\displaystyle
	g(g(Y,n_x^1,n_y^1),n_x^2,n_y^2)=g(Y,n_x^1n_x^2,n_y^1n_y^2)
	$
	\end{lemma}
	\begin{lemma}
	$\displaystyle
	h(h(Y,n_x^1,n_y^1),n_x^2,n_y^2)=h(Y,n_x^1n_x^2,n_y^1n_y^2)
	$
	\end{lemma}
	\begin{lemma}\label{lemma4}
	$\displaystyle
	g(D*W,n_x,n_y)=g(D,n_x,n_y)*h(W,n_x,n_y)
	$
	\end{lemma}
	\begin{lemma}\label{lemma5}
		$\displaystyle
		 h(D*W,n_x,n_y)=h(D,n_x,n_y)*h(W,n_x,n_y)
		$
	\end{lemma}
	The first three lemmas are obvious. Now, we provide the proof of lemma \ref{lemma4} and lemma \ref{lemma5} .\\
	\textbf{Lemma \ref{lemma4}\quad proof}:\\
	Recall that the convolution operator $C=D*W$ means if $D \in {\mathbb R}^{n_{dx} \times n_{dy}}$ and $W \in {\mathbb R}^{n_{wx} \times n_{wy}}$, $C \in {\mathbb R}^{(n_{dx}+n_{wx}+1) \times (n_{dy}+n_{wy}+1) }$ then the $(i,j)^\text{th}$ element is given by
	\begin{align}
	C_{i,j} = \sum_{p=1}^{n_{wx}} \sum_{q=1}^{n_{wy}} D_{i+n_{wx}-p,j+n_{wy}-q} W_{p,q}
	\end{align}
	where 
	\begin{align}
	D_{i,j}=0 \qquad \text{if } i,j<0 \qquad\text{or } i>n_{dx}\qquad \text{or } j>n_{dy}.
	\end{align}
	Let $\hat{C}=g(D*W,n_x,n_y)$, $\hat{D}=g(D,n_x,n_y)$, $\hat{W}=h(W,n_x,n_y)$ and $n_{wx}^{\prime}=(n_{wx}-1)n_x+1$, $n_{wy}^{\prime}=(n_{wy}-1)n_y+1$. We want to prove that $\hat{C}=\hat{D}*\hat{W}$. Deriving elementwise we have
	\begin{align}
	(\hat{D}*\hat{W})_{i^{\prime},j^{\prime}} &= \sum_{p^{\prime}=1}^{n_{wx}^{\prime}} \sum_{q^{\prime}=1}^{n_{wy}^{\prime}} \hat{D}_{i^{\prime}+n_w-p^{\prime},j^{\prime}+n_w-q^{\prime}} \hat{W}_{p^{\prime},q^{\prime}}\nonumber \\
	&=\sum_{p^{\prime}=1}^{n_{wx}^{\prime}} \sum_{q^{\prime}=1}^{n_{wy}^{\prime}} D_{\left\lceil \frac{i^{\prime}+n_{wx}^{\prime}-p^{\prime}}{n_x}\right\rceil,\left\lceil \frac{j^{\prime}+n_{wy}^{\prime}-q^{\prime}}{n_y}\right\rceil} \hat{W}_{p^{\prime},q^{\prime}}\label{proofone}\nonumber\\
	&=\sum_{p=1}^{n_w} \sum_{q=1}^{n_w} D_{\left\lceil \frac{i^{\prime}+(n_{wx}-1)n_x-(p-1)n_x}{n_x}\right\rceil,\left\lceil \frac{j^{\prime}+(n_{wy}-1)n_y-(q-1)n_y}{n_y}\right\rceil}\hat{W}_{(p-1)n_x+1,(q-1)n_x+1}\nonumber\\
	&=\sum_{p=1}^{n_w} \sum_{q=1}^{n_w} D_{\left\lceil \frac{i^{\prime}}{n_x}\right\rceil+n_{wx}-p,\left\lceil \frac{j^{\prime}}{n_y}\right\rceil+n_{wy}-q}W_{p,q}\nonumber\\
	&=C_{\left\lceil \frac{i^{\prime}}{n_x}\right\rceil,\left\lceil \frac{j^{\prime}}{n_y}\right\rceil}\nonumber\\
	&=\hat{C}_{i^{\prime},j^{\prime}}.
	\end{align}
	Since if $p^{\prime}=(p-1)n_x+1$ and $q^{\prime}=(q-1)n_x+1$, then $\hat{W}_{p^{\prime},q^{\prime}}=W_{p,q}$, and if not $\hat{W}_{p^{\prime},q^{\prime}}=0$.\\
	\\
	\textbf{Lemma \ref{lemma5}\quad proof}:\\
	Let $\tilde{C}=h(D*W,n_x,n_y)$, $\tilde{D}=h(D,n_x,n_y)$, $\tilde{W}=h(W,n_x,n_y)$ and $n_{wx}^{\prime}=(n_{wx}-1)n_x+1$, $n_{wy}^{\prime}=(n_{wy}-1)n_y+1$. We want to prove that $\tilde{C}=\tilde{D}*\tilde{W}$. Deriving elementwise we have:
	\begin{align}
	(\tilde{D}*\tilde{W})_{i^{\prime},j^{\prime}} &= \sum_{p^{\prime}=1}^{n_{wx}^{\prime}} \sum_{q^{\prime}=1}^{n_{wy}^{\prime}} \tilde{D}_{i^{\prime}+n_w-p^{\prime},j^{\prime}+n_w-q^{\prime}} \tilde{W}_{p^{\prime},q^{\prime}}\nonumber \\
	&=\sum_{p=1}^{n_w} \sum_{q=1}^{n_w} \tilde{D}_{i^{\prime}+(n_{wx}-p)n_x,j^{\prime}+(n_{wy}-q)n_y}\tilde{W}_{(p-1)n_x+1,(q-1)n_x+1}\nonumber\\
	&=\sum_{p=1}^{n_w} \sum_{q=1}^{n_w} \tilde{D}_{i^{\prime}+(n_{wx}-p)n_x,j^{\prime}+(n_{wy}-q)n_y}W_{p,q}
	\end{align}
	Therefore, when $\frac{i^{\prime}-1}{n_x}$ and $\frac{i^{\prime}-1}{n_y}$ both are integers
	\begin{align}
	(\tilde{D}*\tilde{W})_{i^{\prime},j^{\prime}} &= \sum_{p=1}^{n_w} \sum_{q=1}^{n_w} \tilde{D}_{i^{\prime}+(n_{wx}-p)n_x,j^{\prime}+(n_{wy}-q)n_y}W_{p,q}\nonumber\\
	 &= \sum_{p=1}^{n_w} \sum_{q=1}^{n_w} 
	 D_{\frac{i^{\prime}-1}{n_x}+1+(n_{wx}-p),\frac{j^{\prime}-1}{n_y}+1+(n_{wy}-q)}W_{p,q}\nonumber\\
	 &=(D*W)_{\frac{i^{\prime}-1}{n_x}+1,\frac{j^{\prime}-1}{n_y}+1}\nonumber\\
	 &=C_{\frac{i^{\prime}-1}{n_x}+1,\frac{j^{\prime}-1}{n_y}+1}\nonumber\\
	 &=\tilde{C}_{i^{\prime},j^{\prime}}
	\end{align}
	Otherwise, 	$(\tilde{D}*\tilde{W})_{i^{\prime},j^{\prime}}=0=\tilde{C}_{i^{\prime},j^{\prime}}$. Therefore, $\tilde{C}=\tilde{D}*\tilde{W}$
	\subsection{Layer Collapsing Approximation Method}
	Assume the spatial size of the $l$-th layer dictionary is $N_{dx}^l\times N_{dy}^l$ and the pooling ratio from the $l$-th layer to the $(l+1)$-th layer is $n_x^l\times n_y^l$. 
	Then, the generative model can be formed as:
	\begin{align}
	X^{(n,1)} &= \sum_{k_1=1}^{K_1} D^{(k_1,1)} * f^{-1} (X^{(n,k_1,2)}, Z^{(n,k_1,1)}) + E^{(n)}\\
	X^{(n,k_{l-1},l)} &= \sum_{k_l=1}^{K_l} D^{(k_{l-1},k_l,l)} * f^{-1} (X^{(n,k_l,l+1)}, Z^{(n,k_l,l)})   \qquad  \text{for} \quad l=2,...,L-1\\
	X^{(n,k_{L-1},L)} &= \sum_{k_L=1}^{K_L} D^{(k_{L-1},k_L,L)} * (W^{(n,k_L,L)} \odot Z^{(n,k_L,L)}) 
	\end{align}
	
	The dictionary can be projected down one layer using the following approximation
	\begin{align}
	S^{(n,k_l,l)} &= f^{-1} (X^{(n,k_l,l+1)}, Z^{(n,k_l,l)})\nonumber\\
	&= g(X^{(n,k_l,l+1)}, n_x^l,n_y^l) \odot Z^{(n,k_l,l)}\nonumber\\
	&= g\left(\sum_{k_{l+1}=1}^{K_{l+1}} D^{(k_l,k_{l+1}, l+1)} * S^{(n,k_{l+1},l+1)}, n_x^l,n_y^l\right) \odot Z^{(n,k_l,l)}\nonumber\\
	&= \left[\sum_{k_{l+1}=1}^{K_{l+1}} g(D^{(k_l,k_{l+1},l+1)}, n_x^l,n_y^l) * h(S^{(n,k_{l+1},l+1)}, n_x^l,n_y^l)\right] \odot Z^{(n,k_l,l)}\nonumber\\
	&\approx \sum_{k_{l+1}=1}^{K_{l+1}} 	\left[ g(D^{(k_l,k_{l+1},l+1)}, n_x^l,n_y^l) \odot \hat{Z}^{(n,k_l,l)}\right] * h(S^{(n,k_{l+1},l+1)}, n_x^l,n_y^l)
	\end{align}
	where $\hat{Z}^{(n,k_l,l)}\in \{0,1\}^{N_{dx}^ln_x^l\times N_{dy}^ln_y^l}$. In each $n_x^l\times n_y^l$ block of $\hat{Z}^{(n,k_l,l)}$, there is at most one non-zero element.
	And down two layers by this approximation
	\small{
	\begin{align}
	& S^{(n,k_{l-1},l-1)}= g(\sum_{k_{l}=1}^{K_{l}} D^{(k_{l-1},k_l, l)} * S^{(n,k_{l},l)}, n_x^{l-1},n_y^{l-1}) \odot Z^{(n,k_{l-1},{l-1})}\nonumber\\
	&\approx \sum_{k_{l}=1}^{K_{l}} g(D^{(k_{l-1},k_l, l)}) n_x^{l-1},n_y^{l-1})  * h(S^{(n,k_{l},l)},n_x^{l-1},n_y^{l-1})\nonumber\\
	&= \sum_{k_{l}=1}^{K_{l}} g\left(D^{(k_{l-1},k_l, l)}, n_x^{l-1},n_y^{l-1}\right)  * h\left(\sum_{k_{l+1}=1}^{K_{l+1}} 	\left[ g(D^{(k_l,k_{l+1},l+1)}, n_x^l,n_y^l) \odot \hat{Z}^{(n,k_l,l)}\right] * h(S^{(n,k_{l+1},l+1)}, n_x^l,n_y^l),n_x^{l-1},n_y^{l-1}\right)\nonumber\\
	&= \sum_{k_{l}=1}^{K_{l}} g\left(D^{(k_{l-1},k_l, l)}, n_x^{l-1},n_y^{l-1}\right)  * \sum_{k_{l+1}=1}^{K_{l+1}} \left\{h\left[g(D^{(k_l,k_{l+1},l+1)}, n_x^l,n_y^l) \odot \hat{Z}^{(n,k_l,l)},n_x^{l-1},n_y^{l-1}\right] * h\left(S^{(n,k_{l+1},l+1)}, n_x^{l-1}n_x^l,n_y^{l-1}n_y^l\right)\right\}\nonumber\\
	&= \sum_{k_{l}=1}^{K_{l}}\sum_{k_{l+1}=1}^{K_{l+1}} g\left(D^{(k_{l-1},k_l, l)}, n_x^{l-1},n_y^{l-1}\right)  *  h\left[g(D^{(k_l,k_{l+1},l+1)}, n_x^l,n_y^l) \odot \hat{Z}^{(n,k_l,l)},n_x^{l-1},n_y^{l-1}\right] * h\left(S^{(n,k_{l+1},l+1)}, n_x^{l-1}n_x^l,n_y^{l-1}n_y^l\right)
	\end{align}
	}
	
	Iteratively collapsing all layers we arrive at the reduced model
	\begin{align}
	X^{(n)} &= \sum_{k_1=1}^{K_1} D^{(k_1,1)} * \left[\sum_{k_2=1}^{K_2} \hat{D}^{(k_1,k_2,2)} * \left(\cdots *(\sum_{k_L=1}^{K_L} \hat{D}^{(k_{L-1},k_L,L)} * \hat{S}^{(n,k_L,L)})\right)\right] + E^{(n)}\nonumber\\
	&= \sum_{k_1=1}^{K_1} \sum_{k_2=1}^{K_2} \cdots \sum_{k_L=1}^{K_L} \left(D^{(k_1,1)} *  \hat{D}^{(k_1,k_2,2)}*\cdots* \hat{D}^{(k_{L-1},k_L,L)} * \hat{S}^{(n,k_L,L)} \right) + E^{(n)} \nonumber\\
	&= \sum_{k_L=1}^{K_L} \left(\sum_{k_1=1}^{K_1} \cdots \sum_{k_{L-1}=1}^{K_{L-1}} D^{(k_1,1)} *\hat{D}^{(k_{1},k_2,l)}*\cdots*  \hat{D}^{(k_{L-1},k_L,L)}\right) * \hat{S}^{(n,k_L,L)}  + E^{(n)}\nonumber\\
	&= \sum_{k_L=1}^{K_L} \tilde{D}^{(k_L)} * \hat{S}^{(n,k_L,L)} + E^{(n)}
	\end{align}
	where
	\begin{align}
	\hat{D}^{(k_{l-1},k_l,l)} &= h\left(g(D^{(k_{l-1},k_{l},l)}), \prod_{i=1}^{l-2} n_x^i,\prod_{i=1}^{l-2} n_y^i\right) \odot h\left(\tilde{Z}^{(n,k_{l-1},l)},\prod_{i=1}^{l-2} n_x^i,\prod_{i=1}^{l-2} n_y^i\right), \qquad  \text{for} \quad l=3,...,L-1 \\
	\hat{D}^{(k_{1},k_2,2)}&=g(D^{k_1,k_2,2},n_x^1,n_y^1)\odot \tilde{Z}^{(n,k_{1},1)}\\
	\hat{S}^{(n,k_L,L)} &= h\left(S^{(n,k_{L},L)} , \prod_{l=1}^{L-1}n_x^l,\prod_{l=1}^{L-1}n_y^l\right).
	\end{align}
	and the collapsed (one-layer) dictionary is given by $\{\tilde{D}^{(k_L)}\}_{k_L=1}^{K_L}$.
	
\section{testing \label{Sec:test}}
	When given $M$ testing images $\{X^{(m)}\}_{m=1}^M$, we learn the test feature from:
	\begin{align}
	X^{(m)}=\sum_{k_L=1}^{K_L} \tilde{D}^{k_L}*(W^{(m,k_L)}\odot Z^{(m,k_L}))+E^{(m)} \label{eq:testing}
	\end{align}
	Let $s_m$ be a vector ``unfolded" from $\{S^{(m,k_L)}=W^{(m,k_L)}\odot Z^{m,k_L}\}_{k_L=1}^{K_L}$.  $s_m$ will be sent to a nonlinear
	support vector machine (SVM) with Gaussian kernel, in a one-vs-all multi-class classifier, with classifier parameters tuned via 5-fold cross-validation (no tuning on the deep feature learning).

\section{Algorithm Flow-Chart \label{Sec:flowchart}}

\begin{algorithm}[h!]
\caption{Learning and Classification of the Deep Model}
\begin{algorithmic}[1]
\REQUIRE Input images ${\bf X}$, model layers $L$, dictionary sizes $\{K_l\}_{l=1}^L$, pooling sizes $n_x, n_y$.
\STATE Model Training
\begin{itemize}
\item Bottom-up pretraining with inference equations in Section~\ref{Sec:Gibbs}.
\item Top-down refinement with equations in Section~\ref{Sec:Gibbs}
\end{itemize}
\STATE Project top-layer dictionary into data level with equations in Section~\ref{Sec:Proj}.
\STATE Testing with model in Section~\ref{Sec:test}.
\STATE Features learned by (\ref{eq:testing}) sent to SVM for classification.
\end{algorithmic}
\label{algo:MAPAl}
\end{algorithm}

\section{Figure 1 \label{Sec:Figure1}}
Figure~\ref{fig:max_pool} plots the large version of Figure 1 in the main paper.

\begin{figure}[tbp!]

\centering
\vspace{-3mm}
\includegraphics[width=\textheight, height=\textwidth,angle=90]{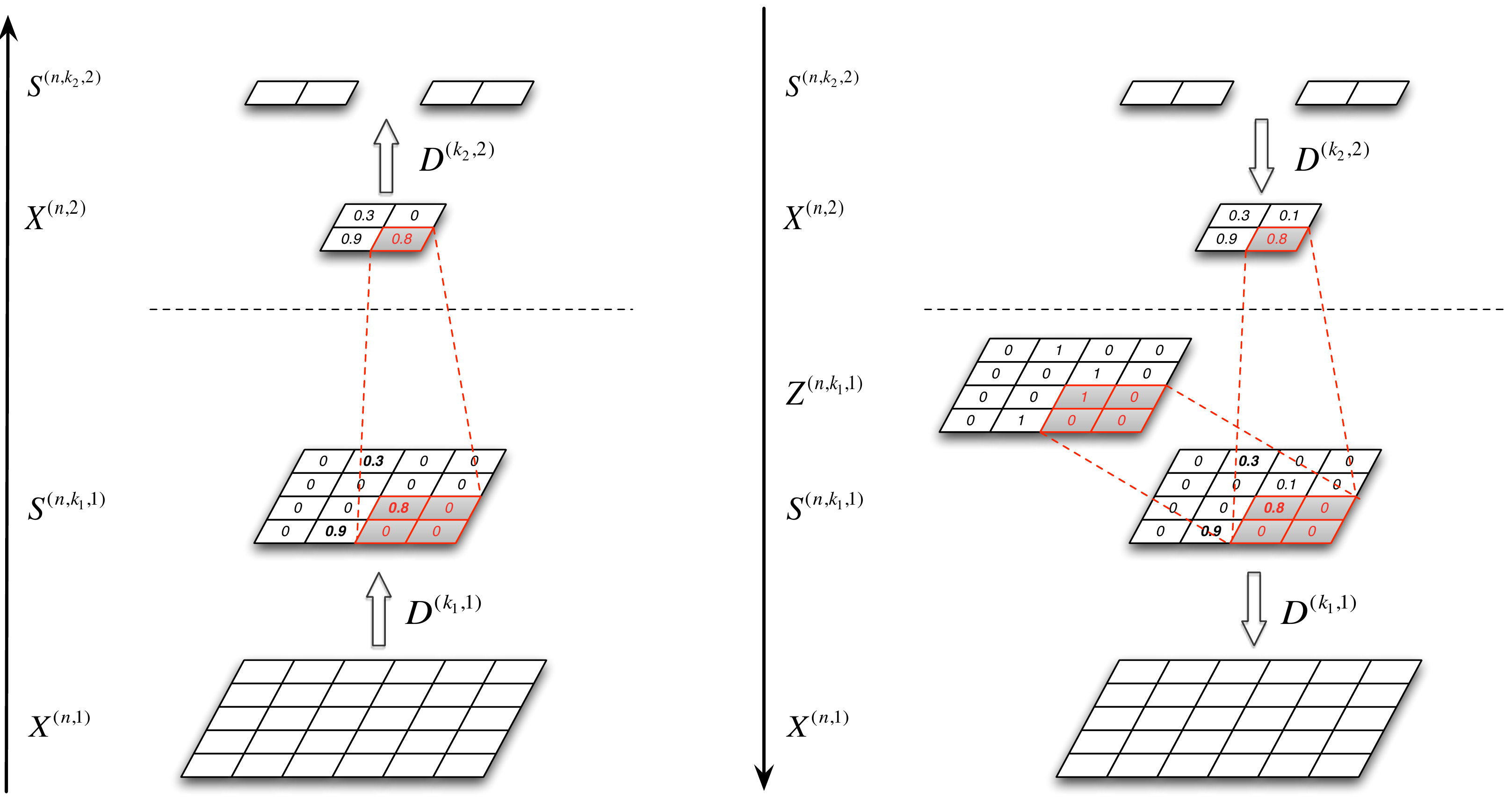}
\vspace{-3mm}
\caption{\small{Schematic of the proposed generative process. Left: bottom-up pretraining, right: top-down refinement.}}
\vspace{-6mm}
\label{fig:max_pool1}
\end{figure}

\section{Generated Images \label{Sec:generate}}
\begin{figure}[htbp!]
	\centering
	\vspace{-3mm}
	\includegraphics[width=\textwidth]{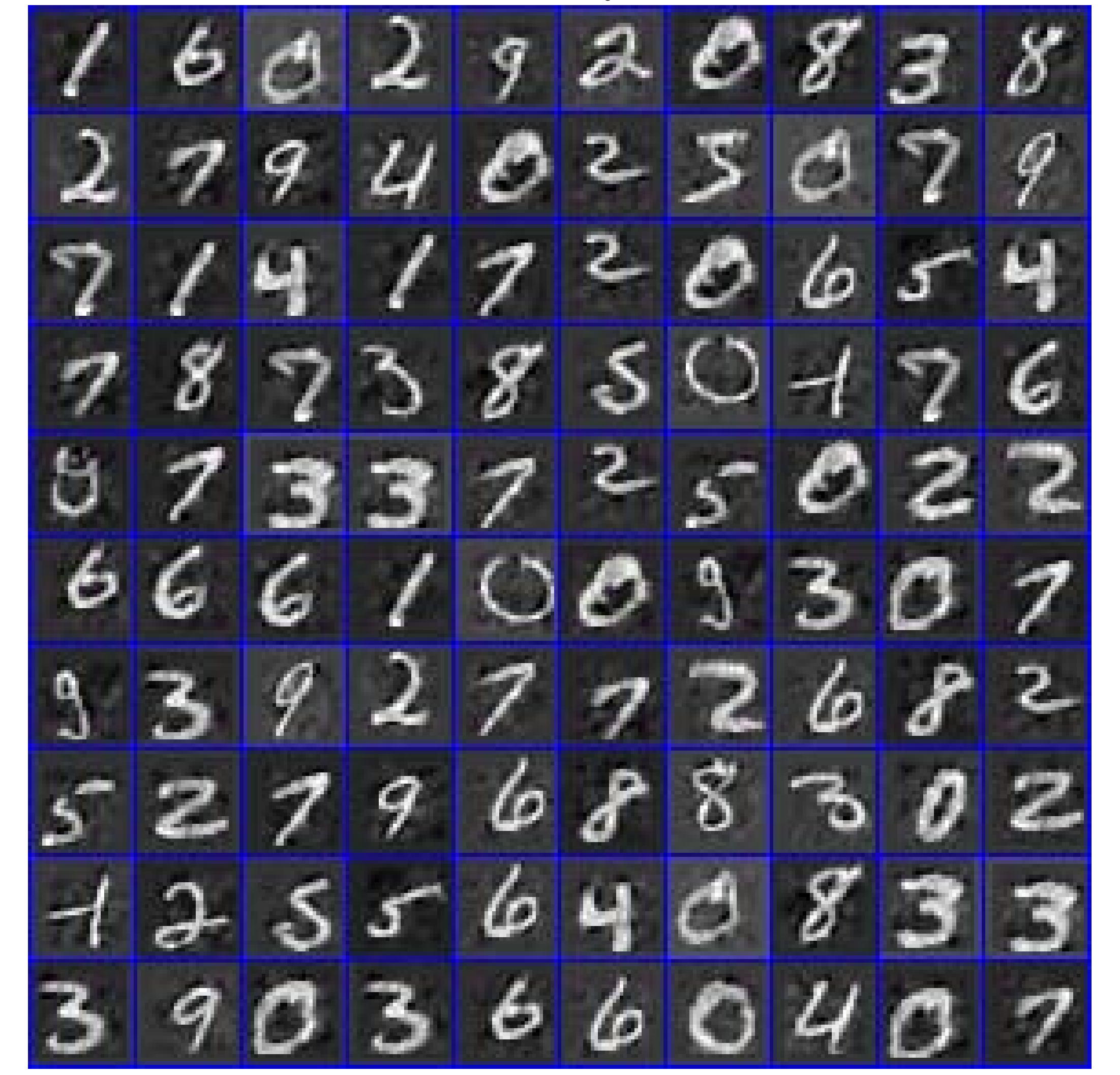}
	\caption{\small{Generated images by randomly drawn dictionary weigths.}}
	\label{Fig:generate}
\end{figure}

We use the trained dictionary by MNIST and randomly generated weights to generate images shown in Figure~\ref{Fig:generate}. It can be seen that, like other neural networks, these images look like digits.

\section{Layer by Layer Classification Results \label{Sec:layerbylayer}}
We also verify the classification results with multiple layer features learned by our model and compare them with the results obtained only by the top layer features. The results are reported in Tables ~\ref{Table:Error_MNIST1} (for MNIST) and \ref{Table:caltech101} (for Caltech 101). 
 	
	\begin{table}[!htbc]
		\caption{\small{Classification Error of MNIST data}}
		\centering
		\begin{tabular}{c|c}
		\hline
		Methods & Test error \\
		\hline
		layer by layer deconvolution & 0.42\%  \\
		One layer deconvolution & 0.42\% \\
		\hline 
		\end{tabular}
		\label{Table:Error_MNIST1}
	\end{table}

\begin{table}[!htbc]
	\caption{\small{Classification Results (accuracy $\%$) of Caltech 101 data}}
	\centering
	\begin{tabular}{c|c c}
	\hline
	Training images per class  & 30\\
		\hline
		layer by layer deconvolution + all layers feature & 84.4$\pm$0.1\%  \\
		One layer deconvolution + one layer feature & 84.3$\pm$0.3\% \\
		\hline 
	\end{tabular}
	\label{Table:caltech101}
\end{table}

\section{More Results \label{Sec:Moreresults}}
\begin{figure}[htbp!]
	\centering
	\vspace{-3mm}
	\includegraphics[width=\textwidth,height = 11cm]{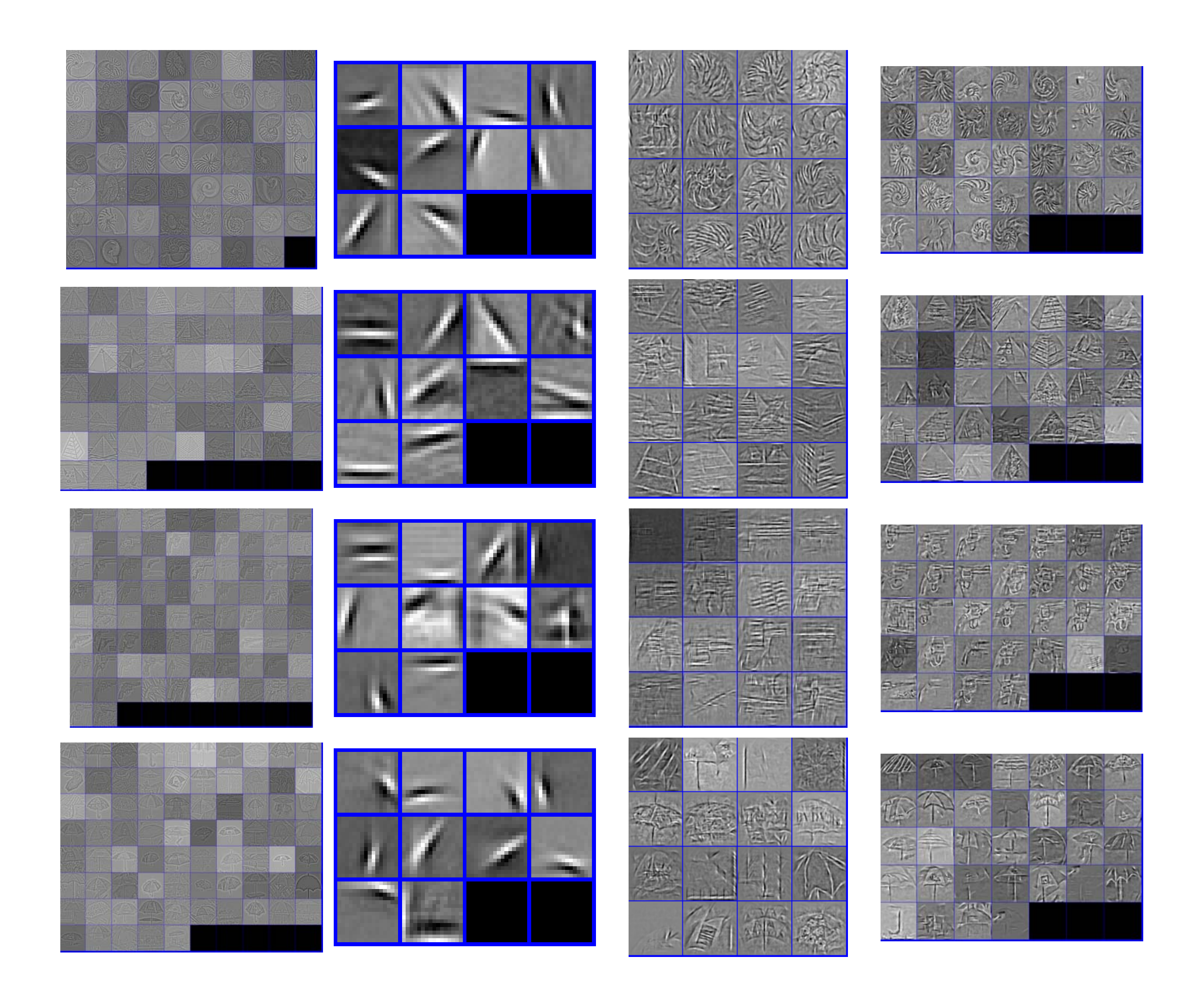}
	\caption{\small{Trained dictionaries per class with each class (from Caltech 101) trained independently. Row 1-4: nautilus, pyramid, revolver, umbrella. Column 1-4: training images after local contrast normalization, layer 1 dictionary, layer 2 dictionary, layer 3 dictionary.}}
	\label{Fig:SepDict1}
\end{figure}
\begin{figure}[htbp!]
	\centering
	\vspace{-3mm}
	\includegraphics[width=\textwidth]{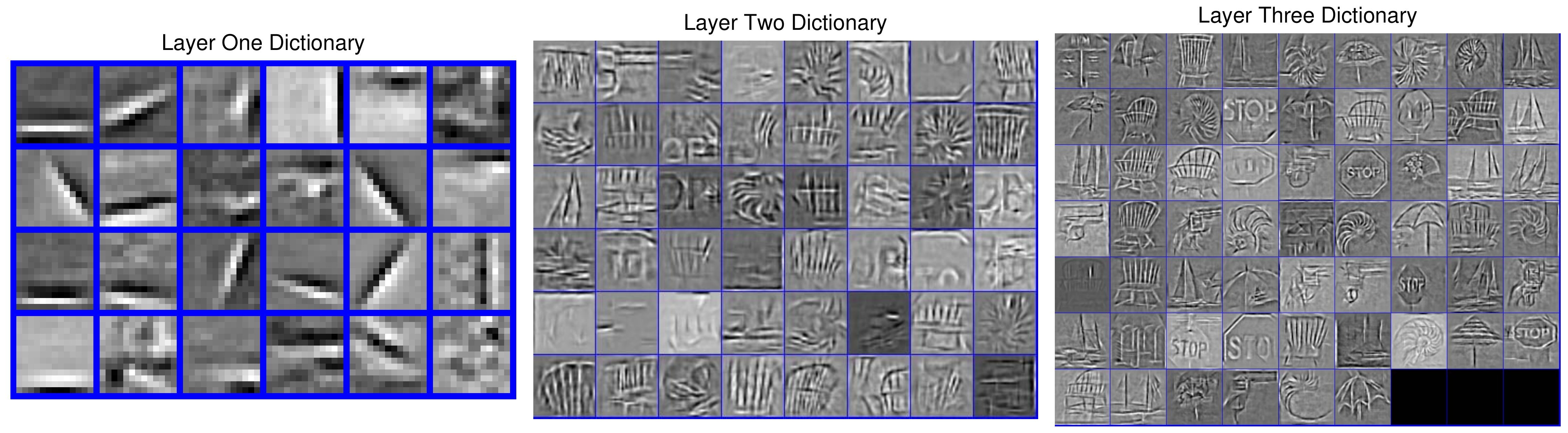}
	\vspace{-2mm}
	\caption{\small{Dictionaries (layer 1 to layer 3) trained by 6 classes (ketch, nautilus, revolver, stop sign, umbrella, windsor chair) images from Caltech 101.}}
	\label{Fig:JointDict}
	\vspace{-3mm}
\end{figure}
Fig.~\ref{Fig:SepDict} plots dictionaries trained per class and four classes are shown. It can be seen the layer one dictionary elements are pretty similar while the upper layer dictionaries vary from each other.
We also performed the model on a joint dataset with multi-class images, with trained dictionary elements shown in Fig.~\ref{Fig:JointDict}. It can be seen that the first layer and second layer dictionaries are more general in this joint training process. For instance, the first layer filters in Fig.~\ref{Fig:JointDict} is an integration of filters of each class in Fig.\ref{Fig:SepDict}.
	
Table~\ref{Table:Error_caltech256} summarizes the classification accuracy of our model and some related models with 15, 30 and 60 training images per class.
So far, only deep convolutional models have provided accuracy greater than $60\%$. 
The best result not using deep convolutional models (using hand-crafted features) is $58\%$ by Bo et al., CVPR 2013 (Bo, L., Ren, X., and Fox, D. Multipath sparse coding using
hierarchical matching pursuit. CVPR 2013.).	
Note that Zeiler \& Fergus, ECCV2014 used a large dataset, namely, the ImageNet, to train the convolutional network, thus to get good results, while we only use the images inside Caltech 256; this same issue exists for the Zeiler \& Fergus, ECCV2014 Caltech 101 results.
	
\begin{table}[htbp!]
		\caption{\small{Classification Accuracy ($\%$) for Caltech 256}}
		\centering
\begin{tabular}{c|ccc}
\hline
\textbf{Training images per class} & \bf{15} & \bf{30} & \bf{60}\\
\hline \hline
Bo et al., CVPR 2013 (Multipath sparse coding)  & 42.7 & 48.0& 58 \\
\hline
Our model & 52.4$\pm$0.3 & 58.3$\pm$0.2   & 65.2$\pm$0.3 \\
\hline
Zeiler \& Fergus, ECCV, 2014 & 65$\pm$0.2 & 70.2$\pm$0.2 & 74.2$\pm$0.3 \\
\hline
\end{tabular}
\label{Table:Error_caltech256}
\end{table}	

\section{Detailed Parameters Used in Experiments \label{Sec:para}}
For MNIST,
a 2-layer model ($K_1 = 32$, $K_2 = 160$) is used. For the first layer, we have 32 dictionary nodes (each with size $8\times 8$) and $32\times 60000$ weights (each with size $21\times 21$). For the second layer, we have 160 dictionary nodes (each with size $6\times 6\times 32$) and $160\times 60000$ weights (each with size $2\times 2$). These parameters and the latent parameters in the model are list in Table~\ref{Table:para_MNIST}.

\begin{table}[htbp!]
		\caption{\small{Parameter number used in MNIST Training}}
		\centering
\begin{tabular}{l|c}
\hline
\textbf{Parameter} & Number\\
\hline \hline
Layer 1, ${\bf D}$  & $32 \times 8 \times 8$ \\
\hline
Layer 1, ${\bf W}$ & $32 \times 21 \times 21 \times 60000$ \\
\hline
Layer 1, ${\bf Z}$ & $32 \times 21 \times 21 \times 60000$ \\
\hline
Layer 2, ${\bf D}$ & $160\times 6\times 6\times 32$\\
\hline
Layer 2, ${\bf W}$ & $160\times 2\times 2\times 60000$\\
\hline
Layer 2, ${\bf Z}$ & $160\times 2\times 2\times 60000$\\
\hline  \hline
\textbf{Latent Parameter} & Number\\
\hline \hline
Layer 1, ${\gamma_e}$ & $60000$ \\
\hline
Layer 1, ${\gamma_d}$ & $32$ \\
\hline
Layer 1, ${\gamma_w}$ & $32\times 21\times 21\times 60000$ \\
\hline
Layer 1, ${\boldsymbol \theta}$  & $32\times 21\times 21 \times 60000$ \\
\hline
Layer 2, ${\gamma_e}$ & $60000$ \\
\hline
Layer 2, ${\gamma_d}$ &  $160\times 32$\\
\hline
Layer 2, ${\gamma_w}$ & $160\times 2\times 2\times 60000$ \\
\hline
Layer 2, ${\boldsymbol \theta}$ & $160\times 2\times 2\times 60000$ \\
\hline
\end{tabular}
\label{Table:para_MNIST}
\end{table}	

For Caltech 101,
in the pretraining step, a 3-layer model ($K_1 = 8$, $K_2 = 12$, $K_3=16$) is used for each category. For the first layer, we have 8 dictionary nodes (each with size $17\times 17$) and $8\times 30$ weights (each with size $112 \times 112$). For the second layer, we have 12 dictionary nodes (each with size $9 \times 9 \times 8$) and $12 \times 30$ weights (each with size $20 \times 20$). For the third layer, we have 16 dictionary nodes (each with size $6 \times 6 \times 12$) and $16 \times 30$ weights (each with size $5 \times 5$). 
These parameters and the latent parameters in the model are list in Table~\ref{Table:para_101}.
In the testing, a single-layer model ($K=1616$) is used. We have $1616\times 6114$ weights (each with size $5\times 5$). 
\begin{table}[htbp!]
		\caption{\small{Parameter number used in Caltech101 Pretraining(Each Category)}}
		\centering
\begin{tabular}{l|c}
\hline
\textbf{Parameter} & Number\\
\hline \hline
Layer 1, ${\bf D}$  & $8 \times 17 \times 17$ \\
\hline
Layer 1, ${\bf W}$ & $8 \times 112\times 112 \times 30$ \\
\hline
Layer 1, ${\bf Z}$ & $8 \times 112\times 112 \times 30$ \\
\hline
Layer 2, ${\bf D}$  & $12 \times 9 \times 9\times 8$ \\
\hline
Layer 2, ${\bf W}$ & $12 \times 20\times 20 \times 30$ \\
\hline
Layer 2, ${\bf Z}$ & $12 \times 20\times 20 \times 30$ \\
\hline
Layer 3, ${\bf D}$  & $16 \times 6 \times 6\times 12$ \\
\hline
Layer 3, ${\bf W}$ & $16 \times 5\times 5 \times 30$ \\
\hline
Layer 3, ${\bf Z}$ & $16 \times 5\times 5 \times 30$ \\
\hline \hline
\textbf{Latent Parameter} & Number\\
\hline \hline
Layer 1, ${\gamma_e}$ & $30$ \\
\hline
Layer 1, ${\gamma_d}$ & $8$ \\
\hline
Layer 1, ${\gamma_w}$ & $8 \times 112\times 112 \times 30$ \\
\hline
Layer 1, ${\boldsymbol \theta}$ & $8 \times 112\times 112 \times 30$ \\
\hline
Layer 2, ${\gamma_e}$ & $30$ \\
\hline
Layer 2, ${\gamma_d}$ & $12\times 8$ \\
\hline
Layer 2, ${\gamma_w}$ & $12 \times 20\times 20 \times 30$ \\
\hline
Layer 2, ${\boldsymbol \theta}$ & $12 \times 20\times 20 \times 30$ \\
\hline
Layer 3, ${\gamma_e}$ & $30$ \\
\hline
Layer 3, ${\gamma_d}$ & $16\times 12$ \\
\hline
Layer 3, ${\gamma_w}$ & $16 \times 5\times 5 \times 30$ \\
\hline
Layer 3, ${\boldsymbol \theta}$  & $16 \times 5\times 5 \times 30$ \\
\hline
\end{tabular}
\label{Table:para_101}
\end{table}

\end{document}